\documentclass[10pt,twocolumn,letterpaper]{article}

\usepackage[pagenumbers]{cvpr} %

\usepackage{times}
\usepackage{epsfig}
\usepackage{graphicx}
\usepackage{amsmath}
\usepackage{subcaption}
\usepackage{amssymb}
\usepackage{hhline}
\usepackage{color, colortbl}
\usepackage{multirow}
\usepackage{times}
\usepackage{epsfig}
\usepackage{graphicx}
\usepackage{arydshln}
\usepackage{amsmath}
\usepackage{amssymb}
\usepackage[utf8]{inputenc}
\usepackage{lipsum} 
\usepackage{booktabs}
\usepackage{xcolor}
\usepackage{makecell}
\usepackage{url}
\usepackage[accsupp]{axessibility}

\usepackage[pagebackref,breaklinks,colorlinks]{hyperref}

\usepackage[capitalize]{cleveref}
\crefname{section}{Sec.}{Secs.}
\Crefname{section}{Section}{Sections}
\Crefname{table}{Table}{Tables}
\crefname{table}{Tab.}{Tabs.}

\def\cvprPaperID{11} %
\def\confName{CVPR}
\def\confYear{2023}

\hypersetup{
urlcolor=blue,
}
\begin{document}

\title{Face Recognition Accuracy Across Demographics:\\Shining a Light Into the Problem}

\author{Haiyu Wu$^{1}$, Vítor Albiero$^{1}$, K. S. Krishnapriya$^{3}$, Michael C. King$^{2}$, Kevin W. Bowyer$^{1}$\\
$^{1}$University of Notre Dame, $^{2}$Florida Institute of Technology, $^{3}$Valdosta State University\\}

\maketitle

\begin{abstract}
We explore varying face recognition accuracy across demographic groups as a phenomenon partly caused by differences in  face illumination.
We observe that for a common operational scenario with controlled image acquisition, there is a large difference in face region brightness between African-American and Caucasian, and also a smaller difference between male and female.
We show that impostor image pairs with both faces under-exposed, or both over-exposed, have an increased false match rate (FMR).
Conversely, image pairs with strongly different face brightness have a decreased similarity measure.
We propose a brightness information metric to measure variation in brightness in the face  and show that face brightness that is too low or too high has reduced information in the face region, providing a cause for the lower accuracy.
Based on this, for operational scenarios with controlled image acquisition, illumination should be adjusted for each individual to obtain appropriate face image brightness.
This is the first work that we are aware of to explore how the level of brightness of the skin region in a pair of face images (rather than a single image) impacts face recognition accuracy, and to evaluate this as a systematic factor causing unequal accuracy across demographics. The code is at \url{https://github.com/HaiyuWu/FaceBrightness}
\end{abstract}

\begin{figure}[t]
    \captionsetup{type=figure}
    \begin{subfigure}[b]{1\linewidth}
        \begin{subfigure}[b]{0.49\linewidth}
          \includegraphics[width=\linewidth]{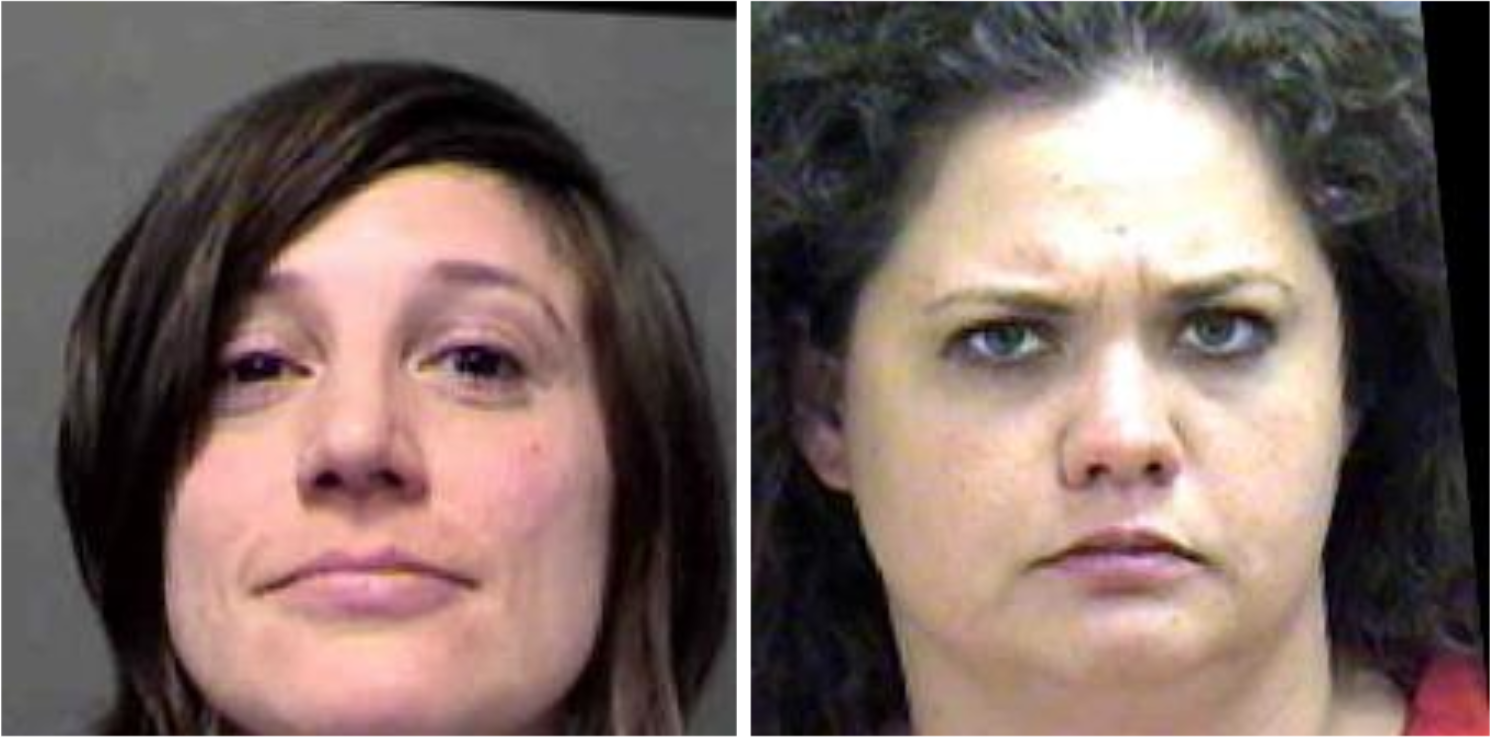}
          \caption{0.0431}
          \label{fig:teaser_image_a}
        \end{subfigure}
        \hfill %
        \begin{subfigure}[b]{0.49\linewidth}
          \includegraphics[width=\linewidth]{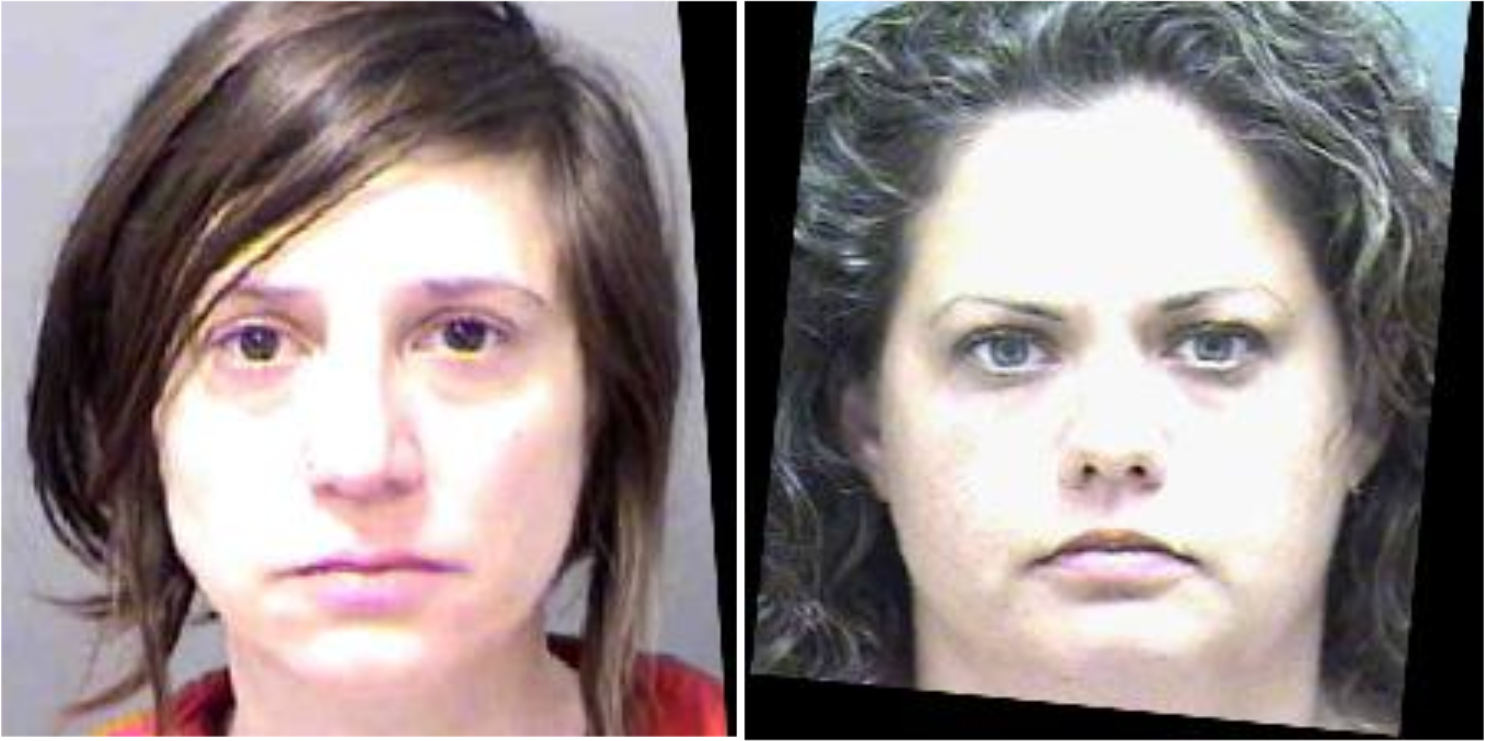}
          \caption{0.2638}
          \label{fig:teaser_image_b}
        \end{subfigure}
        \hfill %
        \begin{subfigure}[b]{0.49\linewidth}
          \includegraphics[width=\linewidth]{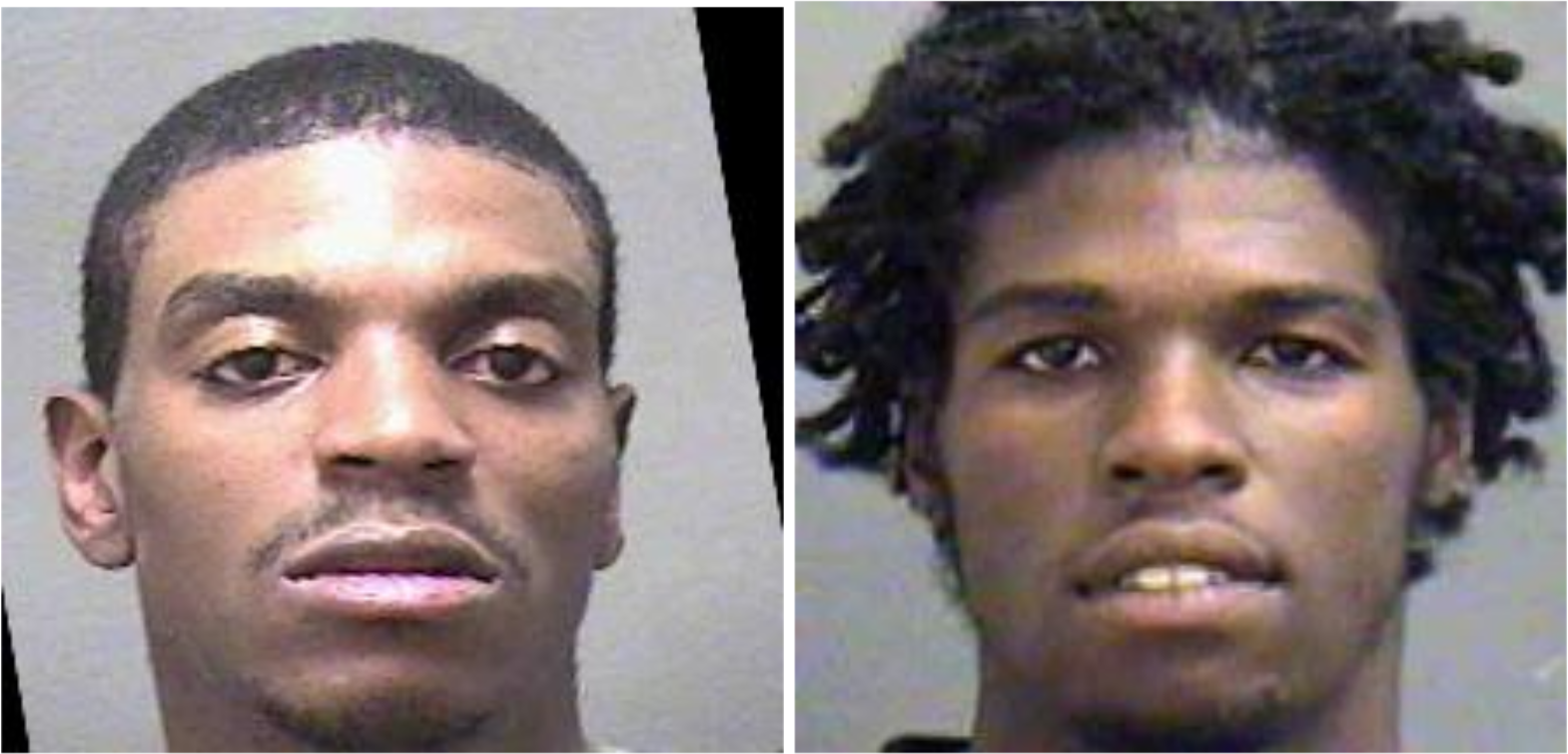}
          \caption{0.0050}
          \label{fig:teaser_image_c}
        \end{subfigure}
        \hfill %
        \begin{subfigure}[b]{0.49\linewidth}
          \includegraphics[width=\linewidth]{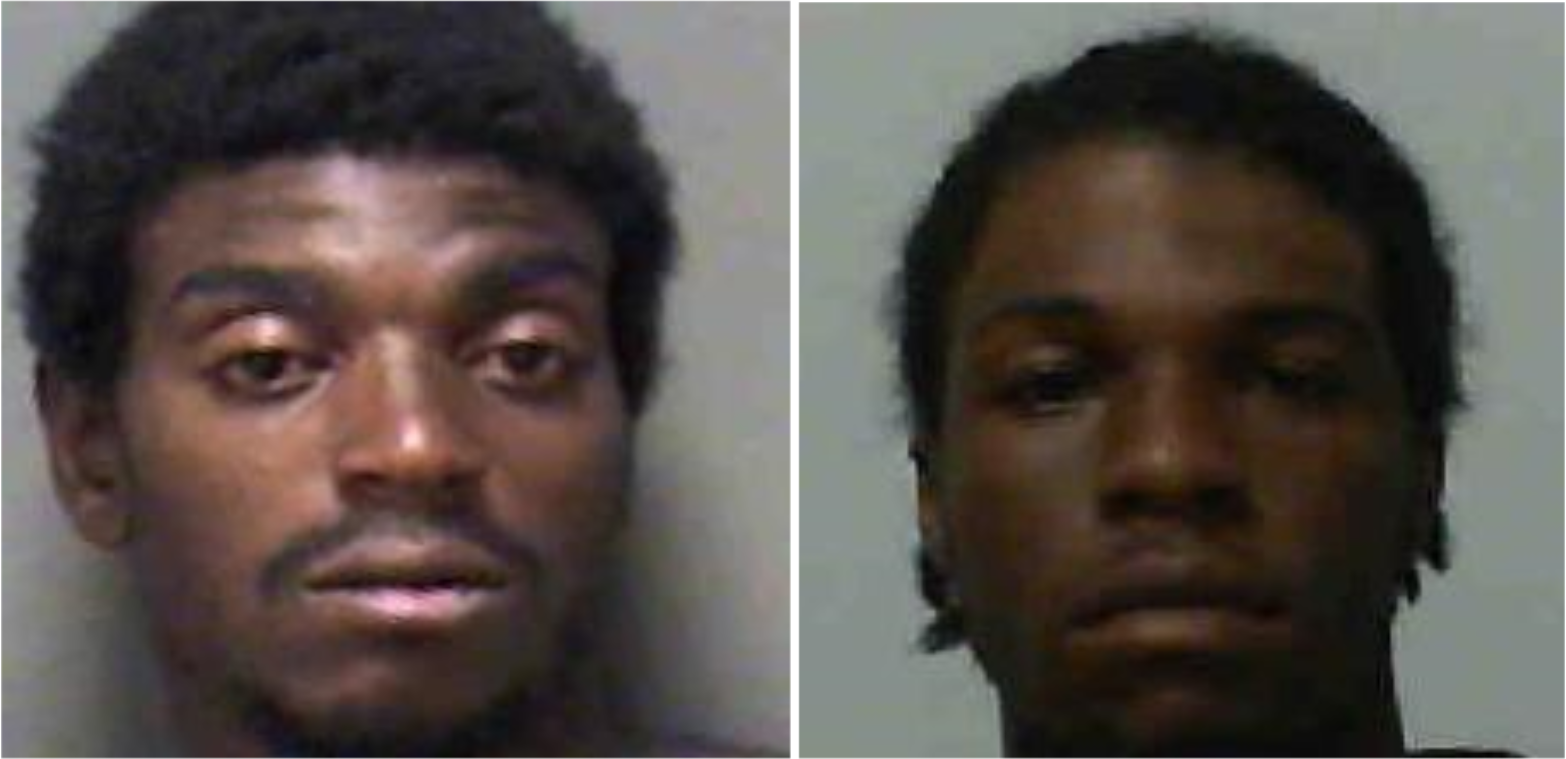}
          \caption{0.3389}
          \label{fig:teaser_image_d}
        \end{subfigure}
    \end{subfigure}
    \vspace{-2em}
    \caption{Impostor image pairs have, on average, higher similarity scores when both images are too bright (``over-exposed") or too dark (``under-exposed"), due to less information available to distinguish between faces. Pairs in (a) and (b) are the same two persons, but images in (a) have appropriate brightness and in (b) are over-exposed, resulting in (b) having a higher similarity score. Similarly, Pairs in (c) and (d) are the same two persons, but images in (c) have appropriate brightness and in (d) are under-exposed, resulting in (d) having a higher similarity score. 
    }
\end{figure}

\section{Introduction}
Widespread adoption of facial recognition technology in recent years has brought the startling realization that accuracy can vary significantly across demographics.
This phenomenon has been widely discussed in news stories \cite{Hoggins2019, Lohr2018, Santow2020, Vincent2019} and research papers \cite{Albiero_2020_BMVC, Cavazos_2021, Drozdowski_2020, FRVT_2019_Part3, krishnapriya2020issues, Terhorst_2021, wu2023logical, bhatta2023gender}.
In response, facial recognition researchers~\cite{Howard_2021,krishnapriya2020issues,Lu_2019, cook2019demographic} have explored the fundamental questions of \textit{why} and \textit{how} recognition accuracy varies across demographics. 
This work approaches the issue of accuracy variation through the lens of image brightness, in the context of controlled image acquisition scenarios such as driver's license, passport, ID card, etc.
The contributions of this work are:
\begin{itemize}
\item 
\vspace{-2mm}
Using an operational dataset acquired under  conditions typical of mugshot, passport or ID card photo, we show that 
African-Americans are more likely to have under-exposed images and Caucasians are there are more likely to have over-exposed images.
(See Section \ref{face_brightness_distributions}.)
\vspace{-2mm}
\item 
We propose a face skin brightness (FSB) metric that avoids regions of the image that are not related to recognition (hair color, sunglasses, ...), to give a more recognition-relevant brightness estimate than a commercial ICAO-compliance SDK.
(See Section \ref{face_skin_brightness}.)
\item 
\vspace{-2mm}
Using ArcFace~\cite{deng2019arcface} and a current commercial-off-the-shelf (COTS) matcher, we show that  
impostor pairs that are ``too dark'' and ``too bright'' can have a FMR twice that of pairs with appropriate brightness.
Also, impostor pairs that have strongly different brightness have a reduced FMR relative to pairs with appropriate brightness.
(See Section \ref{fmr_with_brightness}.)
\item 
\vspace{-2mm}
We provide a target brightness range which supports best accuracy for all demographic (i.e. male/female and Caucasian/African American) for both matchers. It can easily be applied in real life by using the FSB metric.
(See Section \ref{target_brightness}.)
\end{itemize}

\section{Related Work}
\label{literature}
A detailed discussion of face image quality metrics related to illumination appears in a recent survey~\cite{Schlett_2021}. We briefly mention selected related works considering illumination and brightness in face images in different contexts. 

A large body of work was enabled by the Pose, Illumination and Expression (PIE)~\cite{han2013comparative, Sim2003PAMI} and Multi-PIE~\cite{Gross2010IVCJ} datasets.
These datasets acquired images with flashes at various positions around the subject in order to explore the role of lighting direction.
These datasets do not support exploring the effects of the general illumination level.

Beveridge et al~\cite{ Beveridge_2010b} 
concluded that lighting direction can explain significant drops in matching accuracy. 
Han et al~\cite{han2013comparative} surveyed the effect of various illumination preprocessing techniques on PIE and other datasets in terms of improving recognition accuracy for a selection of pre-deep-learning recognition algorithms.
De~Marsico et al~\cite{De_Marsico_2011} proposed an illumination-distortion metric that reflects the degree of uniformity in face illumination, computed as the standard deviation of the brightness of a set of eight face regions. Their metric does not capture the overall level of brightness. 
Abaza et al~\cite{Abaza_2014} proposed a face image quality index based on factors such as contrast, brightness, sharpness, focus and illumination. They used gamma adjustment to create darker and brighter versions of an image. They found that lowered-brightness images have slightly reduced rank-one recognition accuracy and increased-brightness images generally have the same accuracy as the original. 
\textit{\textbf{Their experiment shows higher brightness can increase the recognition rate.}} Later, in Table~\ref{table:FSB}, we show that higher brightness must eventually result in lower accuracy.
Grm et al~\cite{grm2018strengths} created degraded versions of probe images from the LFW dataset~\cite{LFWTech} 
including simulated over-exposure. Their results show that over-bright probe images have lower verification accuracy. 
However, based on Fig. 4 in ~\cite{grm2018strengths}, too much brightness was added, resulting in unrealistic-looking images and possibly affecting their conclusion. Terhörst et al~\cite{terhorst2021pixel} combined a face recognition model and an image-specific quality regression model to generate saliency maps in order to investigate how image quality impacts the face recognition model. They concluded that face regions with high brightness lead to degraded recognition performance.

Some papers use the average brightness divided by the highest brightness value on the face as a quality metric~\cite{long2011near, nasrollahi2009complete}. \textbf{\textit{Taken to the extreme, this metric says that an over-bright image is higher quality.}} 
None of the works mentioned above use their metric to characterize pairs of images compared for recognition.
A very recent work~\cite{huang2022deep} considers the problem of decreased
accuracy for poorly illuminated images and proposes a feature
restoration network. They consider image brightness as a binary attribute of appropriate / dim, and so do not recognize an effect for too-bright images. They also do not have a metric of brightness, and
work with the small Specs on Faces (SoF) dataset having images of just 112 persons. 

Exploring possible bias in recognition accuracy, researchers have investigated whether persons with darker skin tone have decreased accuracy.
Skin tone is typically rated on a scale from 1 (lightest) to 6 (darkest) \cite{krishnapriya2020issues, Lu_2019, ijbb}.
In the Janus program, Mechanical Turk workers rated web-scraped, in-the-wild celebrity images~\cite{ijbb}.
Lu et al~\cite{Lu_2019} analyzed accuracy across skin tone on the Janus dataset and found that the ROC curves degraded from skin tones 1 to 5, but that skin tone 6 had a better ROC than 5.
In other work~\cite{krishnapriya2020issues}, multiple human 
annotators %
rated skin tone for a set of images from the MORPH dataset~\cite{ricanek2006morph}.
The distribution of different skin-tone pairs was compared for the middle and the high-similarity tail of the impostor distribution, with the conclusion that there is no clear evidence of increased FMR for darker skin tone.
In all studies where observers rate the skin tone of a celebrity, 
there is the likelihood that the rating is influenced by recognizing the celebrity and the person's race, rather than the rating being based solely on face brightness in the image.

A 2022 FRVT report \cite{FRVT_2022} perhaps comes the closest to connecting the brightness of an image pair to matching accuracy.
The report states that ``... False negatives are in large part due to one or both photographs being of poor quality, something that can be coupled with demographics. ... dark skin has more challenging dynamic range capture requirements, with underexposed facial regions resulting in reduced information available to the algorithm."
However, this report does not suggest that over-exposure can have the same effect for essentially the same reason.

{\bf Distinctive elements of this work relative to the works described above include the following.}
Most of the above works focus on brightness of a single image rather than an image pair, and do not consider how the effects vary across demographics 
\cite{Abaza_2014, Beveridge_2010b, grm2018strengths, Gross2010IVCJ, De_Marsico_2011, Sim2003PAMI}.
This paper focuses on brightness of the image pair, and on how this affects accuracy across demographics.
There is previous work investigating how manually-rated skin tone, in particular darker skin tones, in a pair of images is associated with recognition accuracy \cite{krishnapriya2020issues, Lu_2019}.
This paper focuses on analyzing automatically-computed face brightness for an image pair,
and how this affects recognition accuracy, across demographics and whether the brightness is too low or too high.

\section{Face Image Brightness Distributions}
There are many web-scraped, in-the-wild, celebrity image datasets: 
MS-Celeb~\cite{MS-Celeb-1M}, WebFace260M\cite{zhu2021webface260m}, VGGFace\cite{parkhi2015deep}
and others.
With web-scraped datasets, one cannot know if any image is brightened, contrast-enhanced or otherwise manipulated.
To study how face region brightness in originally-acquired images affects accuracy, we use the MORPH dataset \cite{morph_site, ricanek2006morph}. 
MORPH images are acquired in controlled conditions typical of mugshot, passport or ID-card photos, including nominally frontal pose, neutral expression, consistent indoor lighting and plain 18\% gray background.
MORPH was assembled from public records, and is widely used in face aging~\cite{Rothe_2018} 
and in study of demographic accuracy variation~\cite{Abdurrahim_2018,Albiero_2022,Drozdowski_2021_ICCVW,krishnapriya2020issues}.
The version of MORPH used contains 127,319 images:
56,245 images of 8,839 African-American males, 24,857 images of 5,929 African-American females, 35,276 images of 8,835 Caucasian males, and 10,941 images of 2,798 Caucasian females. 
Faces were detected and aligned using img2pose~\cite{img2pose}. 

\begin{figure}[t]
    \captionsetup[subfigure]{labelformat=empty}
   \centering
    \begin{subfigure}[b]{1\linewidth}
        \begin{subfigure}[b]{0.24\linewidth}
          \includegraphics[width=\linewidth]{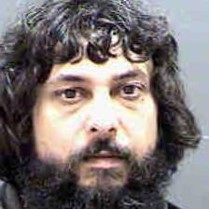}
          \caption{Original}
        \end{subfigure}
        \hfill %
        \begin{subfigure}[b]{0.24\linewidth}
          \includegraphics[width=\linewidth]{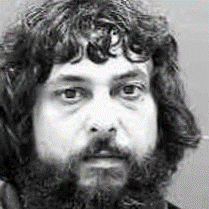}
          \caption{118.37}
        \end{subfigure}
        \hfill %
        \begin{subfigure}[b]{0.24\linewidth}
        
          \includegraphics[width=\linewidth]{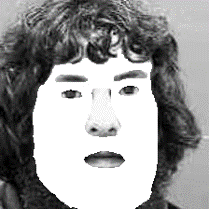}
          \caption{146.37}
        \end{subfigure}
        \hfill %
        \begin{subfigure}[b]{0.24\linewidth}
          \includegraphics[width=\linewidth]{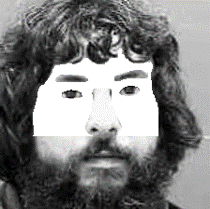}
          \caption{201.69}
        \end{subfigure}
    \end{subfigure}
    \begin{subfigure}[b]{1\linewidth}
        \begin{subfigure}[b]{0.24\linewidth}
          \includegraphics[width=\linewidth]{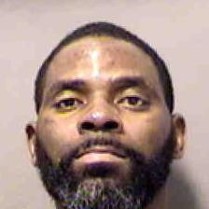}
          \caption{Original}
        \end{subfigure}
        \hfill %
        \begin{subfigure}[b]{0.24\linewidth}
          \includegraphics[width=\linewidth]{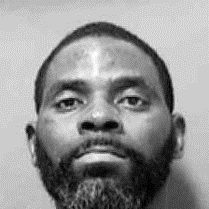}
          \caption{147.13}
        \end{subfigure}
        \hfill %
        \begin{subfigure}[b]{0.24\linewidth}
          \includegraphics[width=\linewidth]{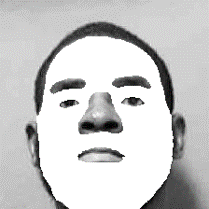}
          \caption{111.46}
        \end{subfigure}
        \hfill %
        \begin{subfigure}[b]{0.24\linewidth}
          \includegraphics[width=\linewidth]{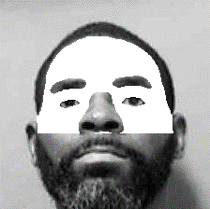}
          \caption{157.66}
        \end{subfigure}
    \end{subfigure}
   \vspace{-6mm}
   \caption{
   Region used for ``Face Skin Brightness'' metric. From left: original color image; using whole image to estimate brightness and its result; using whole face area to estimate brightness and its result; using FSB metric to estimate brightness and its result.}
   \vspace{-5mm}
\label{face_skin_region}
\end{figure}
\begin{figure}[t]
\begin{center}
    \captionsetup[subfigure]{labelformat=empty}
    \begin{subfigure}[b]{1\linewidth}
      \includegraphics[width=\linewidth]{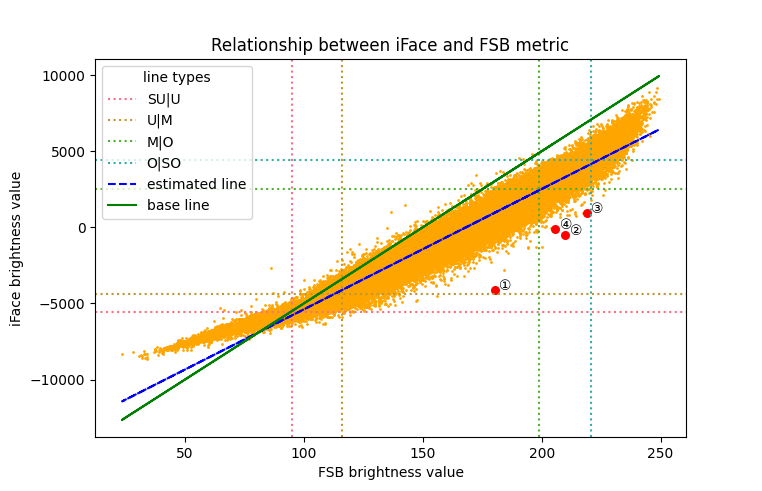}
    \end{subfigure}
    \hfill
    \begin{subfigure}[b]{1\linewidth}
        \begin{subfigure}[b]{0.24\linewidth}
          \includegraphics[width=\linewidth]{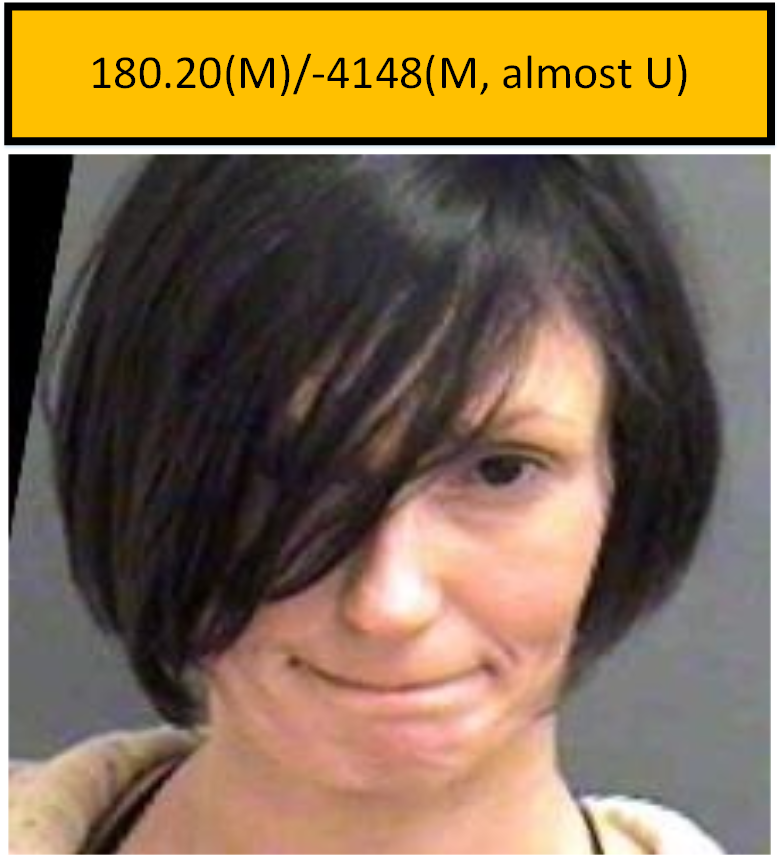}
          \caption{\textcircled{\small 1}}
        \end{subfigure}
        \begin{subfigure}[b]{0.24\linewidth}
          \includegraphics[width=\linewidth]{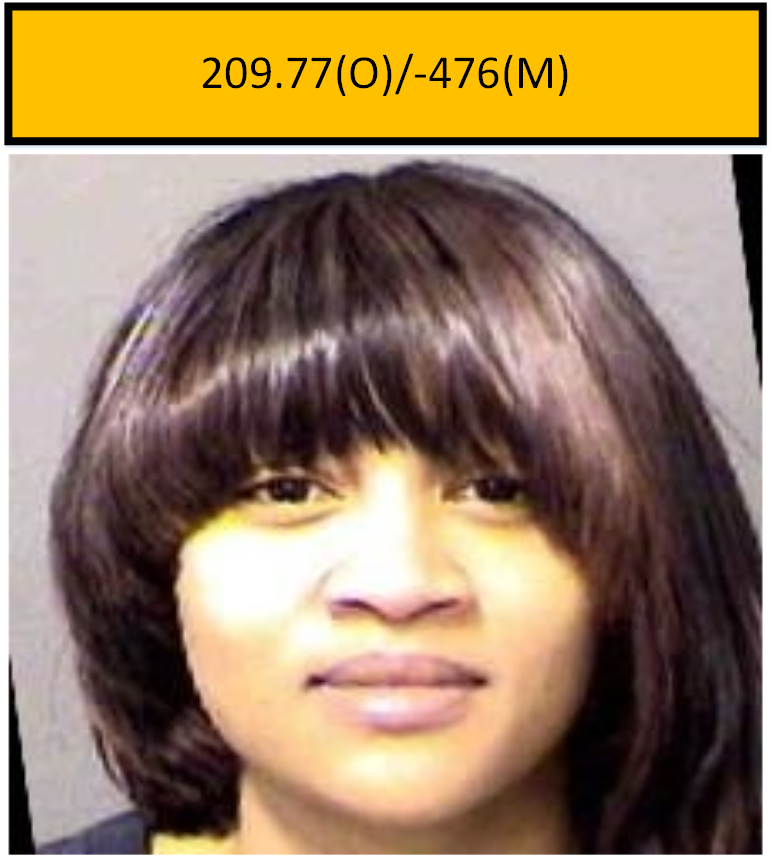}
          \caption{\textcircled{\small 2}}
        \end{subfigure}
        \begin{subfigure}[b]{0.24\linewidth}
          \includegraphics[width=\linewidth]{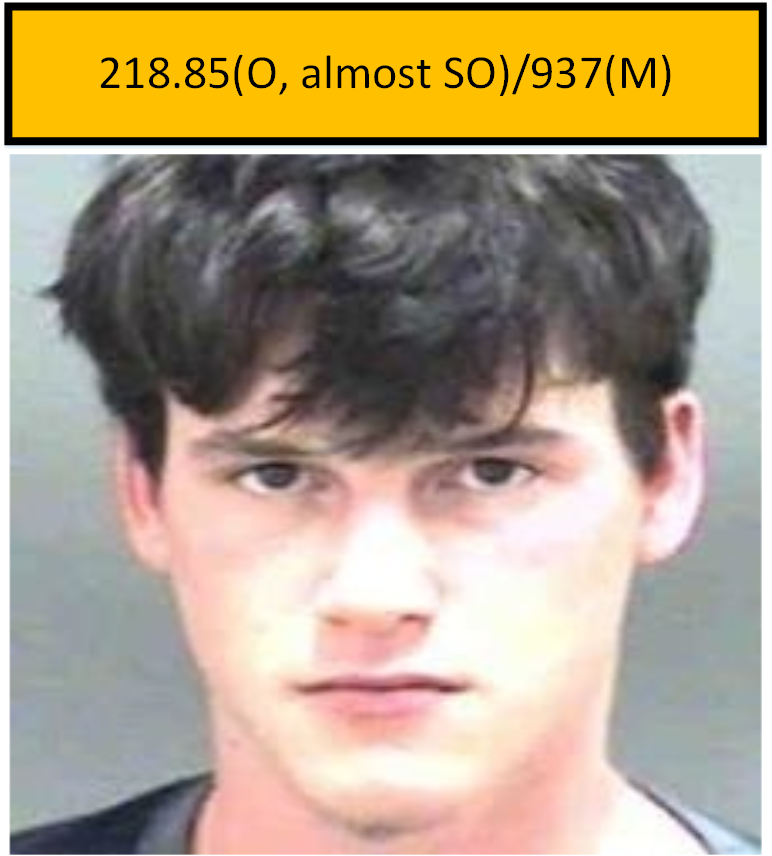}.
          \caption{\textcircled{\small 3}}
        \end{subfigure}
        \begin{subfigure}[b]{0.24\linewidth}
          \includegraphics[width=\linewidth]{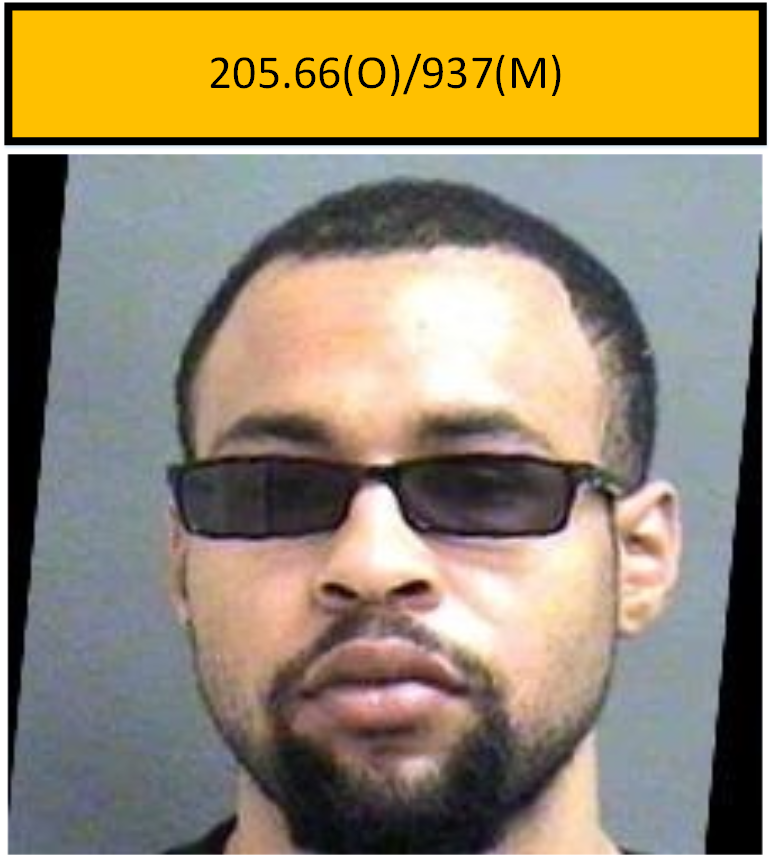}
          \caption{\textcircled{\small 4}}
        \end{subfigure}
    \end{subfigure}
\end{center}
   \vspace{-6mm}
   \caption{Comparison of iFace and FSB brightness values. \textcircled{\small 1}, \textcircled{\small 2}, \textcircled{\small 3}, and \textcircled{\small 4} are the top outliers in each demographic group. These vertical and horizontal dotted lines are the boundaries of each brightness category: \textbf{\textit{Strongly Under-exposed (SU), Under-exposed (U), Middle-exposed (M), Over-exposed (O), and Strongly Over-exposed (SO)}}. The brightness value and brightness group are shown on the top of each example, where the value's format of each person is \textit{$FSB$(group)/$iFace$(group)}.}
   \vspace{-6mm}
\label{fig:brightness_relationship}
\end{figure}

\subsection{Face Skin Brightness Metrics}
\label{face_skin_brightness}

We compare two metrics for estimating brightness of a face image.
One is the metric reported by the iFace SDK~\cite{Iface_SDK, ifacesdk_site}
(version 3.13) for checking compliance with International Civil Aviation Organization (ICAO) standards~\cite{ISO_ICAO,Maltoni_2009}. The other is a metric specifically developed for this work.

ICAO-compliant images have ``appropriate  brightness  and  contrast  that show skin tones naturally''~\cite{icaorules}. 
IFace defines brightness as a ``face attribute for evaluating whether an area of the face is correctly exposed''~\cite{ifacesdk_doc}. Assigned values range from -10,000 (``too dark'') to +10,000 (``too light''), with values near 0 indicating ``OK''. IFace gives decision thresholds of -5,000 for under-exposure and +5,000 for over-exposure.
With the motivation that the face skin region drives face matching accuracy, we introduce a face skin brightness (FSB) metric.
We use the Bilateral Segmentation Network (BiSeNet)~\cite{bisenet_github, yu2018bisenet} to segment a face image into 13 regions, such as face skin, nose, eyes, lips, etc. 
Previous approaches~\cite{abaza2012quality, abaza2014design, cook2019demographic} for estimating the brightness fail to give a correct brightness estimation in the case showing in Figure~\ref{face_skin_region}, because beard and background strongly impact the brightness result. To alleviate the impact from the other factors,
we omit the eyes, eyebrows, and lips from the calculation in order to avoid brightness variation due to sclera, eyelashes/mascara, eyebrows and lipstick.
We also omit the nose region because it often contains specular highlights,
and the skin region below the level of the nose because for men this often contains mustache, or beard.
This resulting region used for our metric is illustrated in the last column of Figure~\ref{face_skin_region}.
FSB is then computed as:
\begin{equation}
\label{eq2} 
    FSB_{image} = \frac{\sum_{h,w \in FS}Image_{gray}(h,w)}{N}
\end{equation}
where $FS$ is the selected face region, $(h, w)$ is the pixel position in $FS$, and $N$ is the number of the pixels in $FS$.
FSB ranges from 0 (darkest) to 255 (brightest).

Figure~\ref{fig:brightness_relationship} compares iFace and FSB brightness for the MORPH images.
iFace estimates relatively higher brightness than FSB for under-exposed images, and relative lower brightness than FSB for the other images.
To better understand how the metrics differ,
Figure~\ref{fig:brightness_relationship} shows the strongest outlier image
for each demographic.
These images are in the upper third of the brightness range of our FSB metric, but significantly lower in the range of the iFace metric.
While we do not know how the iFace metric is calculated, we speculate that hair regions are included in its calculation.

\begin{figure}[t]
\begin{center}
    \captionsetup[subfigure]{labelformat=empty}
    \begin{subfigure}[b]{1\linewidth}
        \begin{subfigure}[b]{0.49\linewidth}
            \centering
            \caption{\textbf{Males}}
            \begin{subfigure}[b]{0.49\linewidth}
              \includegraphics[width=\linewidth]{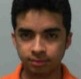}
            \end{subfigure}
            \begin{subfigure}[b]{0.49\linewidth}
              \includegraphics[width=\linewidth]{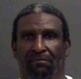}
            \end{subfigure}
        \end{subfigure}
        \begin{subfigure}[b]{0.49\linewidth}
            \centering
            \caption{\textbf{Females}}
            \begin{subfigure}[b]{0.49\linewidth}
              \includegraphics[width=\linewidth]{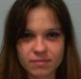}
            \end{subfigure}
            \begin{subfigure}[b]{0.49\linewidth}
              \includegraphics[width=\linewidth]{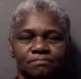}
            \end{subfigure}
        \end{subfigure}
    
    \setcounter{subfigure}{0}
    \caption{(a) Strongly Under-exposed}
    \end{subfigure}
    \begin{subfigure}[b]{1\linewidth}
        \begin{subfigure}[b]{0.24\linewidth}
          \includegraphics[width=\linewidth]{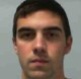}
        \end{subfigure}
        \begin{subfigure}[b]{0.24\linewidth}
          \includegraphics[width=\linewidth]{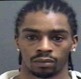}
        \end{subfigure}
        \begin{subfigure}[b]{0.24\linewidth}
          \includegraphics[width=\linewidth]{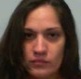}
        \end{subfigure}
        \begin{subfigure}[b]{0.24\linewidth}
          \includegraphics[width=\linewidth]{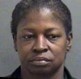}
        \end{subfigure}
    
    \caption{(b) Under-exposed}
    \end{subfigure}
        \begin{subfigure}[b]{1\linewidth}
        \begin{subfigure}[b]{0.24\linewidth}
          \includegraphics[width=\linewidth]{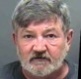}
        \end{subfigure}
        \begin{subfigure}[b]{0.24\linewidth}
          \includegraphics[width=\linewidth]{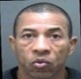}
        \end{subfigure}
        \begin{subfigure}[b]{0.24\linewidth}
          \includegraphics[width=\linewidth]{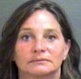}
        \end{subfigure}
        \begin{subfigure}[b]{0.24\linewidth}
          \includegraphics[width=\linewidth]{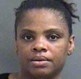}
        \end{subfigure}
    
    \caption{(c) Middle-exposed}
    \end{subfigure}
        \begin{subfigure}[b]{1\linewidth}
        \begin{subfigure}[b]{0.24\linewidth}
          \includegraphics[width=\linewidth]{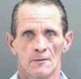}
        \end{subfigure}
        \begin{subfigure}[b]{0.24\linewidth}
          \includegraphics[width=\linewidth]{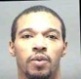}
        \end{subfigure}
        \begin{subfigure}[b]{0.24\linewidth}
          \includegraphics[width=\linewidth]{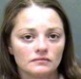}
        \end{subfigure}
        \begin{subfigure}[b]{0.24\linewidth}
          \includegraphics[width=\linewidth]{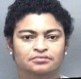}
        \end{subfigure}
    
    \caption{(d) Over-exposed}
    \end{subfigure}
        \begin{subfigure}[b]{1\linewidth}
        \begin{subfigure}[b]{0.24\linewidth}
          \includegraphics[width=\linewidth]{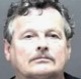}
        \end{subfigure}
        \begin{subfigure}[b]{0.24\linewidth}
          \includegraphics[width=\linewidth]{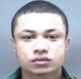}
        \end{subfigure}
        \begin{subfigure}[b]{0.24\linewidth}
          \includegraphics[width=\linewidth]{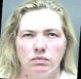}
        \end{subfigure}
        \begin{subfigure}[b]{0.24\linewidth}
          \includegraphics[width=\linewidth]{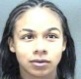}
        \end{subfigure}
    
    \caption{(e) Strongly Over-exposed}
    \end{subfigure}
\end{center}
    \vspace{-6mm}
   \caption{Examples across brightness and demographics.}
   \vspace{-5mm}
\label{fig:demographic_examples}
\end{figure}
\section{Face Brightness Distributions}
\label{face_brightness_distributions}

\begin{figure*}[t]
\centering
    \begin{subfigure}[b]{1\linewidth}
        \begin{tabular}{c:c}
            \begin{subfigure}[b]{0.49\linewidth}
                \begin{subfigure}[b]{0.49\linewidth}
                     \includegraphics[width=\linewidth]{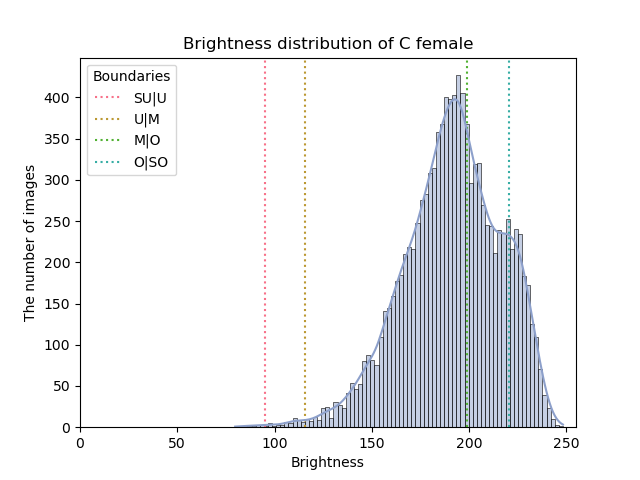}
                \end{subfigure}
                \hfill %
                \begin{subfigure}[b]{0.49\linewidth}
                     \includegraphics[width=\linewidth]{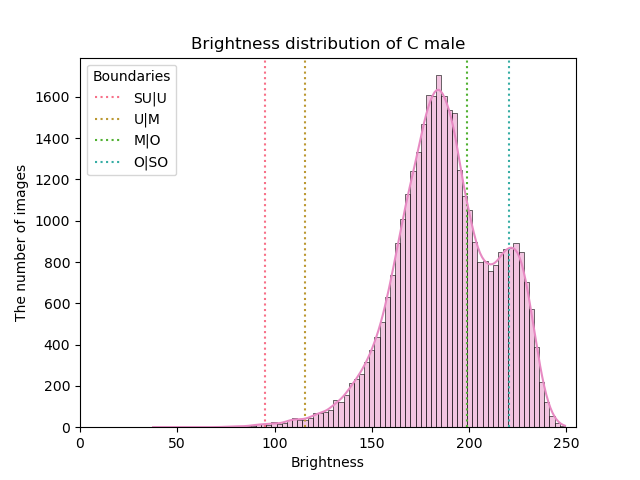}
                \end{subfigure}
            \end{subfigure}
            &
            \begin{subfigure}[b]{0.49\linewidth}
                \begin{subfigure}[b]{0.49\linewidth}
                    \includegraphics[width=\linewidth]{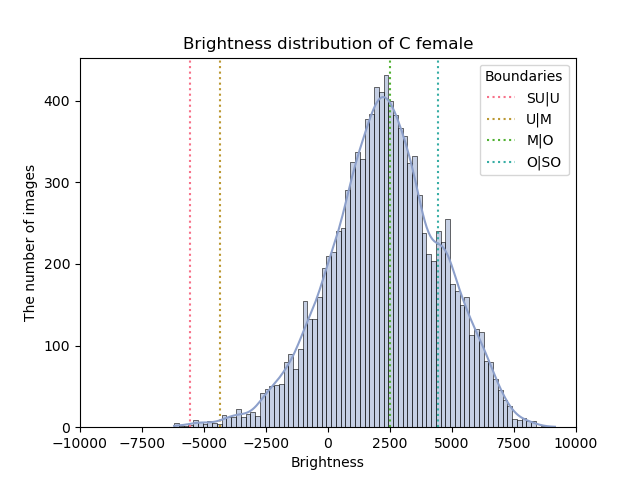}
                \end{subfigure}
                \hfill %
                \begin{subfigure}[b]{0.49\linewidth}
                    \includegraphics[width=\linewidth]{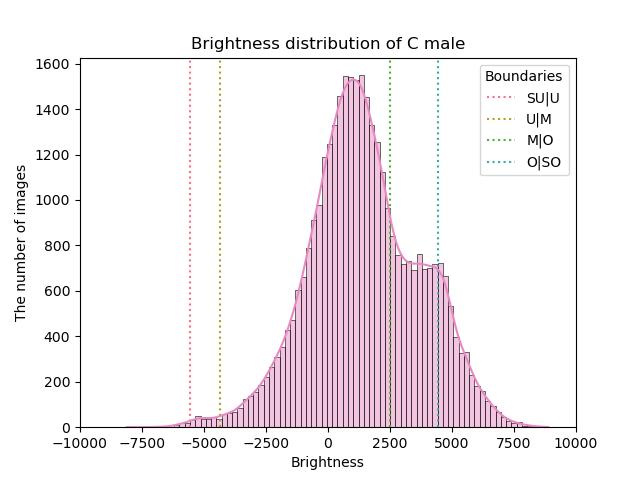}
                \end{subfigure}
            \end{subfigure}
            \\
            \begin{subfigure}[b]{0.49\linewidth}
                \begin{subfigure}[b]{0.49\linewidth}
                     \includegraphics[width=\linewidth]{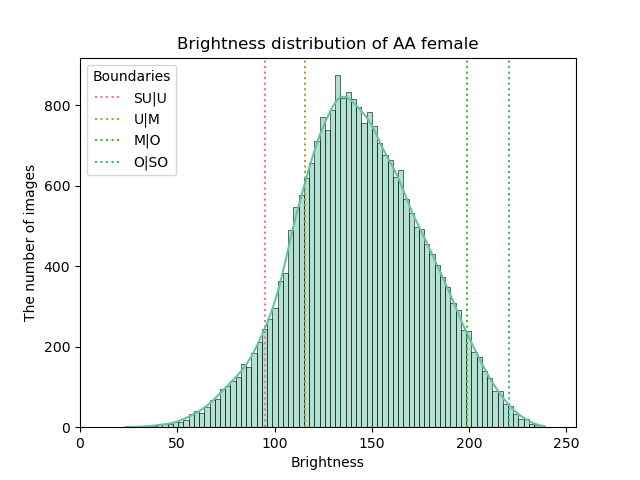}
                \end{subfigure}
                \hfill %
                \begin{subfigure}[b]{0.49\linewidth}
                     \includegraphics[width=\linewidth]{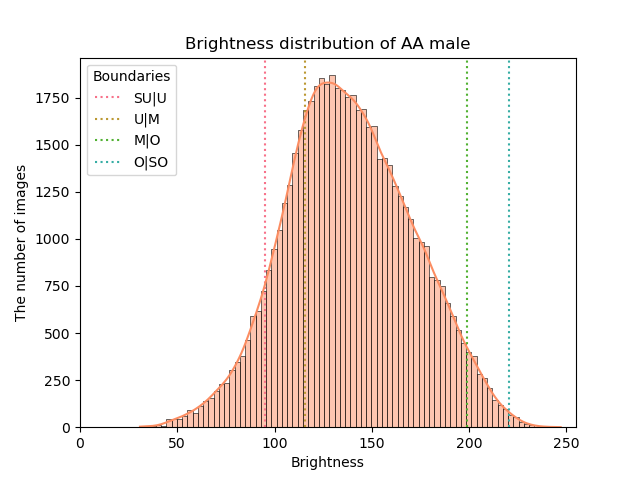}
                \end{subfigure}
                \caption{FSB metric}
            \end{subfigure}
            &
            \begin{subfigure}[b]{0.49\linewidth}
                \begin{subfigure}[b]{0.49\linewidth}
                    \includegraphics[width=\linewidth]{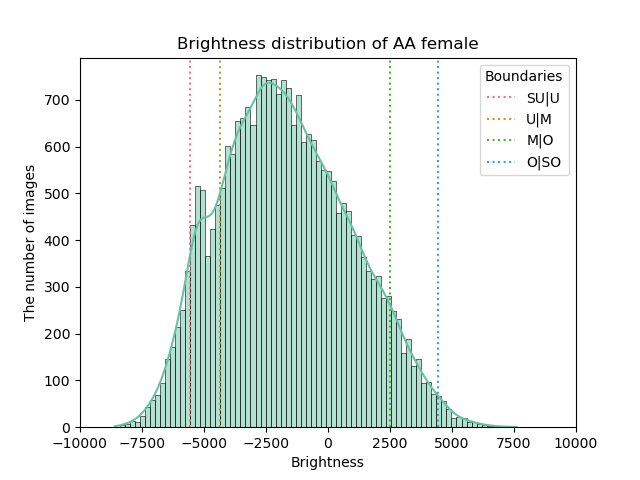}
                \end{subfigure}
                \hfill %
                \begin{subfigure}[b]{0.49\linewidth}
                    \includegraphics[width=\linewidth]{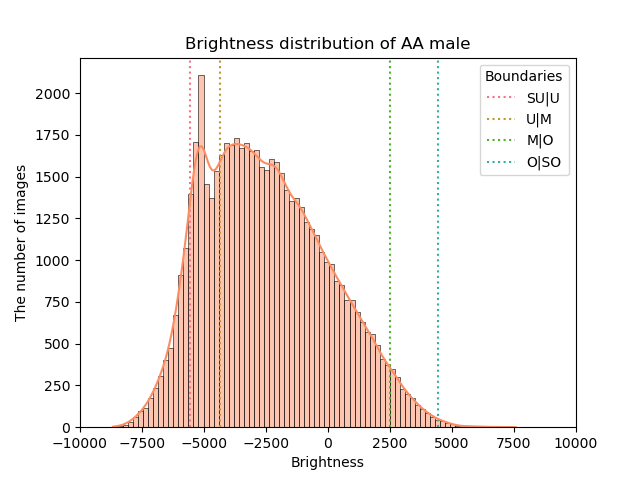}
                \end{subfigure}
                \caption{iFace}
            \end{subfigure}
            \\
        \end{tabular}
    \end{subfigure}
   \caption{Brightness distribution of demographic groups. For each brightness metric, it shows the brightness distributions of Caucasian females, Caucasian males, African-American females, and African-American males. The stacked version is in the Figure 1 of the Supplemental material.}
\label{fig:brightness_distributions}
\end{figure*}

Figure~\ref{fig:brightness_distributions} shows the iFace and FSB distributions for MORPH images, broken out by demographic.
For both iFace and FSB, there is a smaller second peak at higher brightness in the female and male Caucasian distributions.
We speculate this is due to specularities on the forehead in some fraction of the images.
There is a narrow early peak at lower brightness on the iFace (only) distributions for African-American female and male. We speculate this is due to the iFace SDK including hair in the region it uses to estimate brightness.
For both metrics, African-American males have the lowest average brightness (for FSB, mean = 138.7, $\sigma$ = 32.5), followed closely by African-American females (for FSB, mean = 143.5, $\sigma$ = 32.6), then a larger gap to Caucasian males (for FSB, mean = 188.3, $\sigma$ = 25.6), followed closely by Caucasian females (for FSB, mean = 191.4, $\sigma$ = 25).
Note that brightness metrics for African-American females and males have larger standard deviation than for Caucasian females and males.

To analyze how recognition accuracy varies for too-dark or too-bright face skin regions, we divide the brightness distribution for the whole image set into five ranges using the $5\%$, $15\%$, $85\%$, and $95\%$ percentiles.
We refer to the ranges 
as strongly under-exposed (SU), under-exposed (U), middle-exposed (M), over-exposed (O), and strongly over-exposed (SO). 
Figure~\ref{fig:demographic_examples} shows example images for each brightness category and demographic, illustrating 
that the brightness categories are broadly meaningful.
\begin{table*}[t]
\centering
\begin{tabular}{|c|rl||rl||rl||rl|}
\hline
 & \multicolumn{2}{c||}{C Female} & \multicolumn{2}{c||}{C Male} & \multicolumn{2}{c||}{A-A Female} & \multicolumn{2}{c|}{A-A Male} \\ \cline{2-9} 
\multirow{-2}{*}{FSB} & \multicolumn{1}{c|}{Imposters} & \multicolumn{1}{c||}{FMR} & \multicolumn{1}{c|}{Imposters} & \multicolumn{1}{c||}{FMR} & \multicolumn{1}{c|}{Imposters} & \multicolumn{1}{c||}{FMR} & \multicolumn{1}{c|}{Imposters} & \multicolumn{1}{c|}{FMR} \\ \hline
 & \multicolumn{1}{r|}{{\color[HTML]{A6A6A6} }} & {\color[HTML]{A6A6A6} 0.0000} & \multicolumn{1}{r|}{{\color[HTML]{A6A6A6} }} & {\color[HTML]{A6A6A6} 0.0606} & \multicolumn{1}{r|}{} & 0.5188 & \multicolumn{1}{r|}{} & 0.0462 \\ \cline{3-3} \cline{5-5} \cline{7-7} \cline{9-9} 
\multirow{-2}{*}{(SU,SU)} & \multicolumn{1}{r|}{\multirow{-2}{*}{{\color[HTML]{A6A6A6} 90}}} & {\color[HTML]{A6A6A6} 0.0000} & \multicolumn{1}{r|}{\multirow{-2}{*}{{\color[HTML]{A6A6A6} 1,650}}} & {\color[HTML]{A6A6A6} 0.0000} & \multicolumn{1}{r|}{\multirow{-2}{*}{1,329,858}} & 0.0285 & \multicolumn{1}{r|}{\multirow{-2}{*}{10,860,733}} & 0.0051 \\ \hline
 & \multicolumn{1}{r|}{{\color[HTML]{A6A6A6} }} & {\color[HTML]{A6A6A6} 0.0818} & \multicolumn{1}{r|}{{\color[HTML]{A6A6A6} }} & {\color[HTML]{A6A6A6} 0.0000} & \multicolumn{1}{r|}{} & 0.4872 & \multicolumn{1}{r|}{} & 0.0440 \\ \cline{3-3} \cline{5-5} \cline{7-7} \cline{9-9} 
\multirow{-2}{*}{(U,U)} & \multicolumn{1}{r|}{\multirow{-2}{*}{{\color[HTML]{A6A6A6} 1,222}}} & {\color[HTML]{A6A6A6} 0.0000} & \multicolumn{1}{r|}{\multirow{-2}{*}{{\color[HTML]{A6A6A6} 22,938}}} & {\color[HTML]{A6A6A6} 0.0000} & \multicolumn{1}{r|}{\multirow{-2}{*}{5,154,158}} & 0.0339 & \multicolumn{1}{r|}{\multirow{-2}{*}{42,810,922}} & 0.0046 \\ \hline
 & \multicolumn{1}{r|}{} & 0.0359 & \multicolumn{1}{r|}{} & 0.0096 & \multicolumn{1}{r|}{} & 0.3273 & \multicolumn{1}{r|}{} & 0.0374 \\ \cline{3-3} \cline{5-5} \cline{7-7} \cline{9-9} 
\multirow{-2}{*}{(M,M)} & \multicolumn{1}{r|}{\multirow{-2}{*}{21,922,011}} & 0.0528 & \multicolumn{1}{r|}{\multirow{-2}{*}{273,558,170}} & 0.0096 & \multicolumn{1}{r|}{\multirow{-2}{*}{176,625,162}} & 0.0263 & \multicolumn{1}{r|}{\multirow{-2}{*}{812,147,804}} & 0.0034 \\ \hline
 & \multicolumn{1}{r|}{} & 0.0550 & \multicolumn{1}{r|}{} & 0.0121 & \multicolumn{1}{r|}{{\color[HTML]{A6A6A6} }} & {\color[HTML]{A6A6A6} 0.2987} & \multicolumn{1}{r|}{} & 0.0919 \\ \cline{3-3} \cline{5-5} \cline{7-7} \cline{9-9} 
\multirow{-2}{*}{(O,O)} & \multicolumn{1}{r|}{\multirow{-2}{*}{3,851,462}} & 0.0698 & \multicolumn{1}{r|}{\multirow{-2}{*}{24,685,360}} & 0.0123 & \multicolumn{1}{r|}{\multirow{-2}{*}{{\color[HTML]{A6A6A6} 582,234}}} & {\color[HTML]{A6A6A6} 0.0477} & \multicolumn{1}{r|}{\multirow{-2}{*}{1,703,422}} & 0.0067 \\ \hline
 & \multicolumn{1}{r|}{} & 0.1014 & \multicolumn{1}{r|}{} & 0.0255 & \multicolumn{1}{r|}{{\color[HTML]{A6A6A6} }} & {\color[HTML]{A6A6A6} 0.4324} & \multicolumn{1}{r|}{{\color[HTML]{A6A6A6} }} & {\color[HTML]{A6A6A6} 0.3136} \\ \cline{3-3} \cline{5-5} \cline{7-7} \cline{9-9} 
\multirow{-2}{*}{(SO,SO)} & \multicolumn{1}{r|}{\multirow{-2}{*}{1,087,394}} & 0.0921 & \multicolumn{1}{r|}{\multirow{-2}{*}{10,487,792}} & 0.0166 & \multicolumn{1}{r|}{\multirow{-2}{*}{{\color[HTML]{A6A6A6} 9,020}}} & {\color[HTML]{A6A6A6} 0.1441} & \multicolumn{1}{r|}{\multirow{-2}{*}{{\color[HTML]{A6A6A6} 14,986}}} & {\color[HTML]{A6A6A6} 0.0267} \\ \hline \hline
 & \multicolumn{1}{r|}{{\color[HTML]{A6A6A6} }} & {\color[HTML]{A6A6A6} 0.0000} & \multicolumn{1}{r|}{{\color[HTML]{A6A6A6} }} & {\color[HTML]{A6A6A6} 0.0161} & \multicolumn{1}{r|}{} & 0.4919 & \multicolumn{1}{r|}{} & 0.0438 \\ \cline{3-3} \cline{5-5} \cline{7-7} \cline{9-9} 
\multirow{-2}{*}{(SU,U)} & \multicolumn{1}{r|}{\multirow{-2}{*}{{\color[HTML]{A6A6A6} 699}}} & {\color[HTML]{A6A6A6} 0.0000} & \multicolumn{1}{r|}{\multirow{-2}{*}{{\color[HTML]{A6A6A6} 12,447}}} & {\color[HTML]{A6A6A6} 0.0080} & \multicolumn{1}{r|}{\multirow{-2}{*}{5,238,898}} & 0.0030 & \multicolumn{1}{r|}{\multirow{-2}{*}{43,133,866}} & 0.0048 \\ \hline
 & \multicolumn{1}{r|}{{\color[HTML]{A6A6A6} }} & {\color[HTML]{A6A6A6} 0.0166} & \multicolumn{1}{r|}{} & 0.0077 & \multicolumn{1}{r|}{} & 0.3521 & \multicolumn{1}{r|}{} & 0.0338 \\ \cline{3-3} \cline{5-5} \cline{7-7} \cline{9-9} 
\multirow{-2}{*}{(U,M)} & \multicolumn{1}{r|}{\multirow{-2}{*}{{\color[HTML]{A6A6A6} 331,078}}} & {\color[HTML]{A6A6A6} 0.0099} & \multicolumn{1}{r|}{\multirow{-2}{*}{5,027,776}} & 0.0017 & \multicolumn{1}{r|}{\multirow{-2}{*}{60,364,176}} & 0.0241 & \multicolumn{1}{r|}{\multirow{-2}{*}{372,979,809}} & 0.0034 \\ \hline
 & \multicolumn{1}{r|}{} & 0.0386 & \multicolumn{1}{r|}{} & 0.0094 & \multicolumn{1}{r|}{} & 0.2047 & \multicolumn{1}{r|}{} & 0.0366 \\ \cline{3-3} \cline{5-5} \cline{7-7} \cline{9-9} 
\multirow{-2}{*}{(M,O)} & \multicolumn{1}{r|}{\multirow{-2}{*}{18,385,534}} & 0.0553 & \multicolumn{1}{r|}{\multirow{-2}{*}{164,378,978}} & 0.0099 & \multicolumn{1}{r|}{\multirow{-2}{*}{20,297,481}} & 0.0253 & \multicolumn{1}{r|}{\multirow{-2}{*}{74,433,988}} & 0.0033 \\ \hline
 & \multicolumn{1}{r|}{} & 0.0575 & \multicolumn{1}{r|}{} & 0.0137 & \multicolumn{1}{r|}{{\color[HTML]{A6A6A6} }} & {\color[HTML]{A6A6A6} 0.2842} & \multicolumn{1}{r|}{{\color[HTML]{A6A6A6} }} & {\color[HTML]{A6A6A6} 0.1068} \\ \cline{3-3} \cline{5-5} \cline{7-7} \cline{9-9} 
\multirow{-2}{*}{(O,SO)} & \multicolumn{1}{r|}{\multirow{-2}{*}{4,096,167}} & 0.0754 & \multicolumn{1}{r|}{\multirow{-2}{*}{32,189,078}} & 0.0126 & \multicolumn{1}{r|}{\multirow{-2}{*}{{\color[HTML]{A6A6A6} 145,656}}} & {\color[HTML]{A6A6A6} 0.0665} & \multicolumn{1}{r|}{\multirow{-2}{*}{{\color[HTML]{A6A6A6} 321,113}}} & {\color[HTML]{A6A6A6} 0.0106} \\ \hline \hline
 & \multicolumn{1}{r|}{{\color[HTML]{A6A6A6} }} & {\color[HTML]{A6A6A6} 0.0183} & \multicolumn{1}{r|}{} & 0.0071 & \multicolumn{1}{r|}{} & 0.3341 & \multicolumn{1}{r|}{} & 0.0309 \\ \cline{3-3} \cline{5-5} \cline{7-7} \cline{9-9} 
\multirow{-2}{*}{(SU,M)} & \multicolumn{1}{r|}{\multirow{-2}{*}{{\color[HTML]{A6A6A6} 92,668}}} & {\color[HTML]{A6A6A6} 0.0053} & \multicolumn{1}{r|}{\multirow{-2}{*}{1,356,380}} & 0.0010 & \multicolumn{1}{r|}{\multirow{-2}{*}{30,671,103}} & 0.0198 & \multicolumn{1}{r|}{\multirow{-2}{*}{187,881,881}} & 0.0031 \\ \hline
 & \multicolumn{1}{r|}{{\color[HTML]{A6A6A6} }} & {\color[HTML]{A6A6A6} 0.0166} & \multicolumn{1}{r|}{} & 0.0052 & \multicolumn{1}{r|}{} & 0.1362 & \multicolumn{1}{r|}{} & 0.0168 \\ \cline{3-3} \cline{5-5} \cline{7-7} \cline{9-9} 
\multirow{-2}{*}{(U,O)} & \multicolumn{1}{r|}{\multirow{-2}{*}{{\color[HTML]{A6A6A6} 138,810}}} & {\color[HTML]{A6A6A6} 0.0079} & \multicolumn{1}{r|}{\multirow{-2}{*}{1,510,801}} & 0.0019 & \multicolumn{1}{r|}{\multirow{-2}{*}{3,468,769}} & 0.0129 & \multicolumn{1}{r|}{\multirow{-2}{*}{17,093,291}} & 0.0016 \\ \hline
 & \multicolumn{1}{r|}{} & 0.0372 & \multicolumn{1}{r|}{} & 0.0091 & \multicolumn{1}{r|}{} & 0.1123 & \multicolumn{1}{r|}{} & 0.0216 \\ \cline{3-3} \cline{5-5} \cline{7-7} \cline{9-9} 
\multirow{-2}{*}{(M,SO)} & \multicolumn{1}{r|}{\multirow{-2}{*}{9,773,081}} & 0.0524 & \multicolumn{1}{r|}{\multirow{-2}{*}{107,153,315}} & 0.0098 & \multicolumn{1}{r|}{\multirow{-2}{*}{2,537,284}} & 0.0227 & \multicolumn{1}{r|}{\multirow{-2}{*}{7,012,583}} & 0.0026 \\ \hline 
\end{tabular}
\vspace{-2mm}
\caption{FMR across impostor pair brightness categories for all demographic groups. For the FMR of each category, top number is ArcFace model, bottom number is COTS-D. The categories with less than 1M impostors are set in gray.}
\label{table:FSB}
\end{table*}
Note that under- and over-exposure has strong demographic correlation.
African-American females and males have substantial numbers of image pairs with (U,U) and (SU,SU) brightness, but very few with (O,O) or (SO,SO).
The reverse applies to Caucasian females and males, but have substantial numbers of (O,O) and (SO,SO) but very few (U,U) or (SU,SU).

\section{Accuracy Varies By Image Pair Brightness}
\label{fmr_with_brightness}

To check if accuracy variation based on image pair brightness is matcher-dependent, we use two matchers: ArcFace~\cite{deng2019arcface,insightface}, and a recent commercial matcher, COTS-D (not named due to license restriction).
The pattern of results across brightness categories and demographics is similar for the two matchers.
Because the two metrics give similar results and FSB is less affected by non-skin elements, this section presents results only for FSB-based categories.

For each matcher, we compute the impostor distribution separately for each demographic, and then 
select the threshold corresponding to a 1-in-10,000 FMR for the Caucasian male demographic as the threshold for all demographics.
This method of setting the FMR threshold follows the recent NIST report on demographic effects in face recognition accuracy~\cite{FRVT_2019_Part3}.
Also, this method makes the cross-demographic differences in FMR more readily apparent. 

Table~\ref{table:FSB} summarizes the FMRs across the brightness categories, for both matchers and the four demographic groups.
There are five categories of similar-brightness impostor pairs: (SU,SU), (U,U), (M,M), (O,O) and (SO-SO); four categories of pairs with images one brightness category apart; (SU,U), (U,M), (M,O) and (O,SO);
and so on.
The number of image pairs varies greatly across brightness category and demographic.
Categories with less than 1M image pairs are shown in gray in Table~\ref{table:FSB}, to indicate lower statistical support.
Our comments focus primarily on categories with larger numbers of image pairs.
We comment first on how FMR varies with brightness of the image pairs, then on FMR differences across demographic groups, and last on differences between matchers.

One general result for both matchers and for all four demographics is that (M,M) brightness image pairs have lower FMR compared to either (U,U) and (SU,SU) pairs, or to (O,O) and (SO,SO) pairs. 
For example, for Caucasian male, going from brightness category (M,M) to (SO,SO) increases ArcFace FMR from 0.0096 to 0.0255 (166\%), and increases COTS-D FMR from 0.0096 to 0.0166 (73\%). 
For Caucasian female, going from brightness (M,M) to (SO,SO) increases ArcFace FMR from 0.0359 to 0.1014 (183\%), and increases COTS-D FMR from 0.0528 to 0.0921 (74\%). 
For African-American male, going from brightness (M,M) to (SU,SU) increases ArcFace FMR from 0.0374 to 0.0462 (24\%), and increases COTS-D FMR from 0.0034 to 0.0051 (50\%). 
For African-American female, going from brightness (M,M) to (U,U) increases ArcFace FMR from 0.3273 to 0.4872 (49\%), and increases COTS-D FMR from 0.0263 to 0.0339 (29\%). 
The (SU,U) brightness pairs show similar increases in FMR for the African-American demographics, as do the (O,SO) brightness pairs for the Caucasian demographics.
In general, similar brightness image pairs that are too dark or too bright have increased FMR.

\begin{figure*}[h]
\begin{subfigure}[b]{1\linewidth}
    \begin{subfigure}[b]{0.5\linewidth}
      \includegraphics[width=\linewidth]{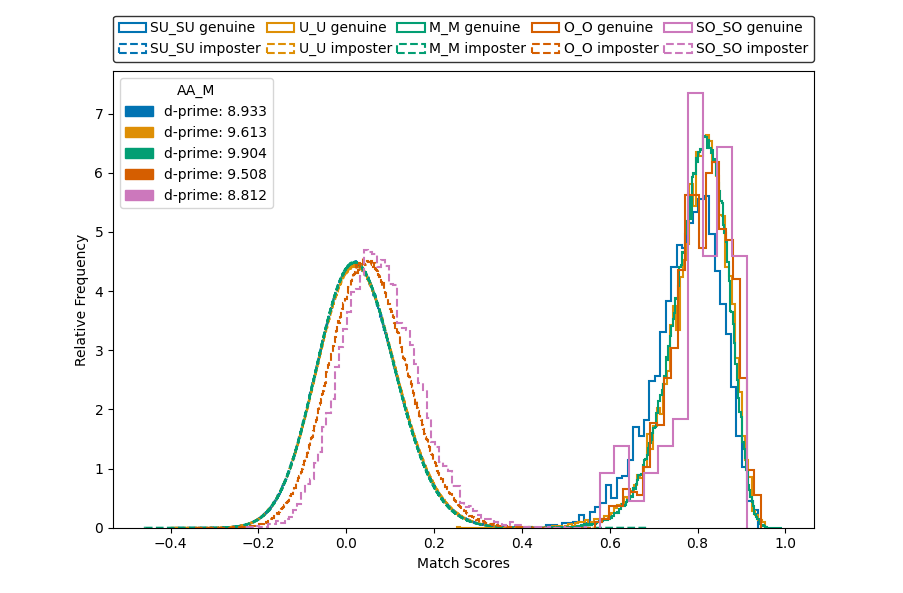}
    \end{subfigure}
    \hfill %
    \begin{subfigure}[b]{0.48\linewidth}
      \includegraphics[width=\linewidth]{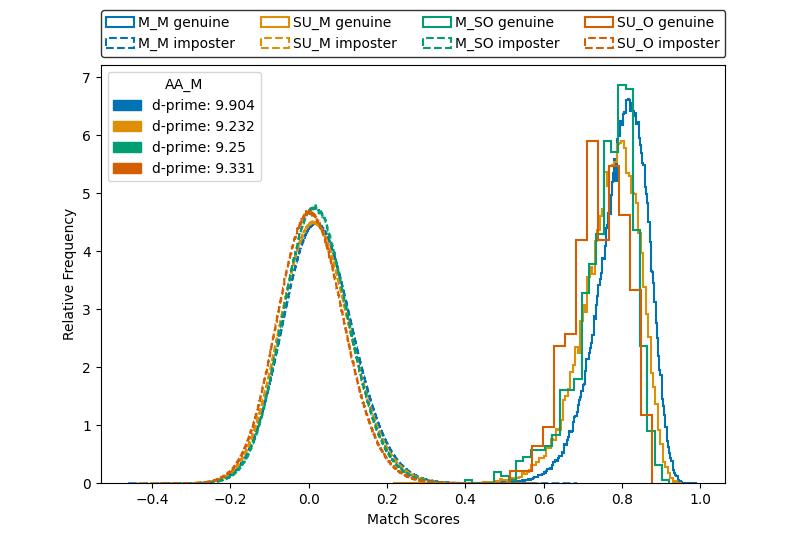}
    \end{subfigure}
\end{subfigure}
  \caption{Genuine and impostor distributions for selected pair brightness categories for African-American Male. The African American male demographic has the most data over the broadest range of brightness categories, enabling better estimates of the genuine distribution than the other demographics. Distributions in the plot on left show that too-dark or too-bright image pairs have both worse impostor and worse genuine distributions.  Distributions in the plot on the right show that image pairs with strong brightness difference have better impostor and worse genuine distributions.}
 \vspace{-5mm}
\label{fig:AAM_imposter_genuine_distribution}
\end{figure*}

Another general result is that image pairs that differ significantly in brightness have a {\it decreased} FMR.
For example, for Caucasian male, (U,O) pairs have an ArcFace FMR of 0.0052 and a COTS-D FMR of 0.0019, {\it decreases} of 45\% and 80\%, respectively, from the FMR for (M,M) pairs.
For African-American male, (U,O) pairs have an FMR of 0.0168 and a COTS-D FMR of 0.0016, {\it decreases} of 55\% and 53\%, respectively, from the FMR for (M,M) pairs.
For African-American female, (U,O) pairs have an FMR of 0.0168 and a COTS-D FMR of 0.0016, {\it decreases} of 58\% and 51\% from the FMR for (M,M) pairs.

\begin{figure*}[h]
\begin{center}
  \begin{subfigure}[b]{1\linewidth}
    \begin{subfigure}[b]{0.33\linewidth}
      \includegraphics[width=\linewidth]{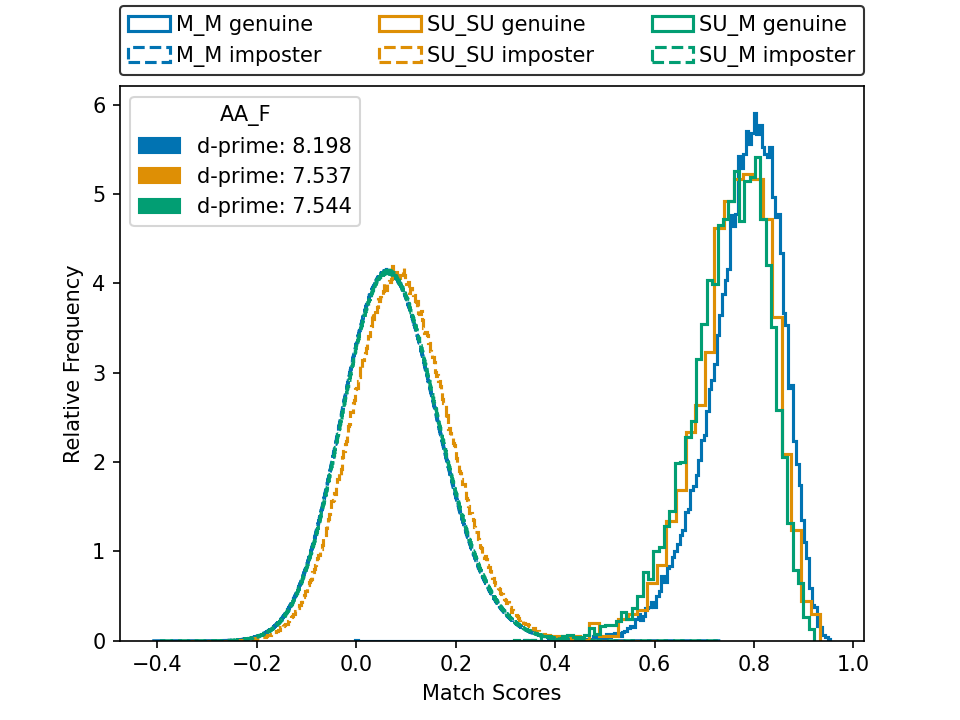}
    \end{subfigure}
    \hfill %
    \begin{subfigure}[b]{0.33\linewidth}
      \includegraphics[width=\linewidth]{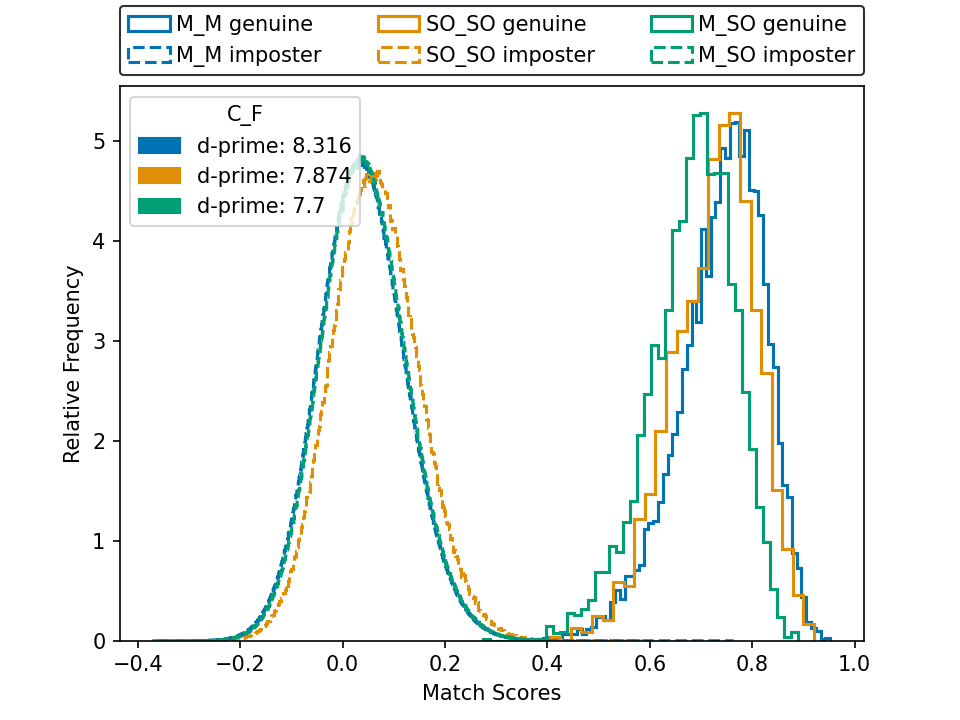}
    \end{subfigure}
    \hfill %
    \begin{subfigure}[b]{0.33\linewidth}
      \includegraphics[width=\linewidth]{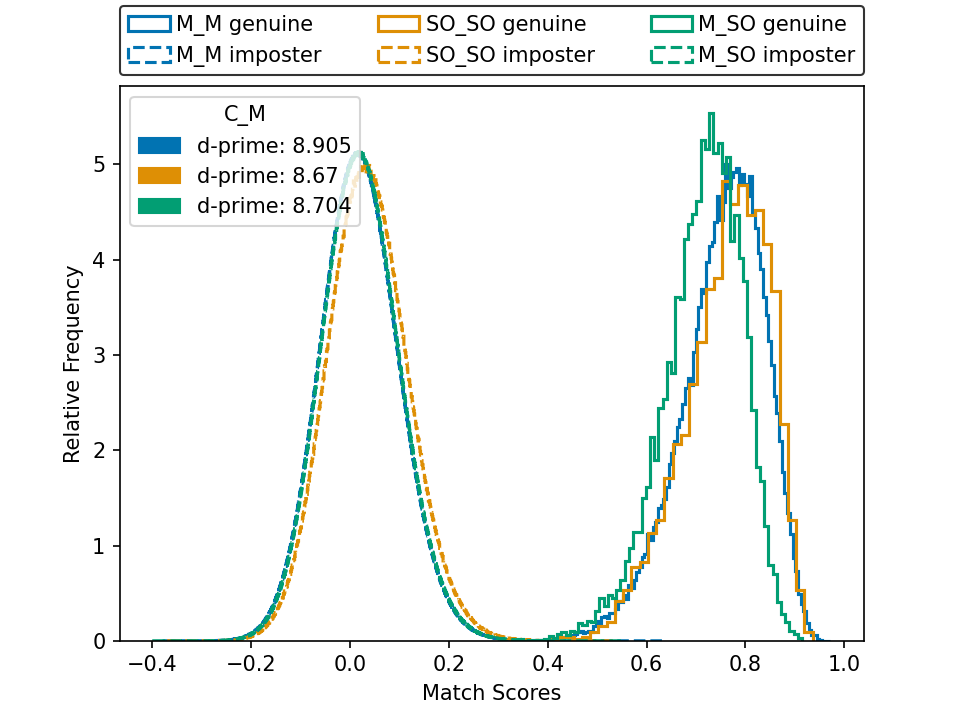}
    \end{subfigure}
\end{subfigure}
\end{center}
\vspace{-6mm}
  \caption{Selected impostor and genuine distributions for African American female, Caucasian male, and Caucasian female.}
\label{fig:AAF_CM_CF_imposter_genuine_distribution}
\end{figure*}

The larger effects of image pair brightness can be seen in the impostor and genuine distributions in Figure~\ref{fig:AAM_imposter_genuine_distribution}.
The distributions in the left plot
show that image pairs that are strongly under- or over-exposed have an impostor distribution that is shifted to increased similarity scores, meaning a higher FMR, and also a lower d-prime for separation of impostor and genuine distributions.
The distributions in the right plot 
show that image pairs that have strongly different brightness have an impostor distribution shifted to lower similarity, meaning a lower FMR, but also a genuine distribution shifted even further to lower similarity, meaning a higher FNMR and resulting in a decreased d-prime.
This pattern of accuracy variation is reinforced by the impostor and genuine distributions for other demographics shown in Figure~\ref{fig:AAF_CM_CF_imposter_genuine_distribution}.
Image pairs of extreme low or high brightness result in an impostor distribution shifted to higher similarity, and image pairs of strongly different brightness result in a genuine distribution shifted to lower similarity.

\begin{table*}[t]
\centering
\begin{tabular}{|c|ccl|ccl|ccl|ccl|}
\hline
\multirow{2}{*}{FSB} & \multicolumn{3}{c|}{C Female} & \multicolumn{3}{c|}{C Male} & \multicolumn{3}{c|}{A-A Female} & \multicolumn{3}{c|}{A-A Male} \\ \cline{2-13} 
 & \multicolumn{1}{c|}{BIM} & \multicolumn{1}{c|}{FMR} & $d'$ & \multicolumn{1}{c|}{BIM} & \multicolumn{1}{c|}{FMR} & $d'$ & \multicolumn{1}{c|}{BIM} & \multicolumn{1}{c|}{FMR} & $d'$ & \multicolumn{1}{c|}{BIM} & \multicolumn{1}{c|}{FMR} & $d'$ \\ \hline
(SU,SU) & \multicolumn{1}{c|}{23.72} & \multicolumn{1}{c|}{ - } &\multicolumn{1}{c|}{ - } & \multicolumn{1}{c|}{23.19} & \multicolumn{1}{c|}{ - } & \multicolumn{1}{c|}{ - } & \multicolumn{1}{c|}{24.52} & \multicolumn{1}{c|}{0.5188} & 7.5 & \multicolumn{1}{c|}{24.49} & \multicolumn{1}{c|}{0.0462} & 8.9 \\ \hline
(U,U) & \multicolumn{1}{c|}{26.70} & \multicolumn{1}{c|}{ - } &\multicolumn{1}{c|}{ - } & \multicolumn{1}{c|}{27.05} & \multicolumn{1}{c|}{ - } &\multicolumn{1}{c|}{ - } & \multicolumn{1}{c|}{29.61} & \multicolumn{1}{c|}{0.4872} & 7.9 & \multicolumn{1}{c|}{29.69} & \multicolumn{1}{c|}{0.044} & 9.6 \\ \hline
(M,M) & \multicolumn{1}{c|}{\textbf{35.98}} & \multicolumn{1}{c|}{\textbf{0.0359}} & \textbf{8.3} & \multicolumn{1}{c|}{\textbf{34.69}} & \multicolumn{1}{c|}{\textbf{0.0096}} & \textbf{8.9} & \multicolumn{1}{c|}{\textbf{34.53}} & \multicolumn{1}{c|}{0.3273} & \textbf{8.2} & \multicolumn{1}{c|}{\textbf{34.97}} & \multicolumn{1}{c|}{\textbf{0.0374}} & \textbf{9.9} \\ \hline
(O,O) & \multicolumn{1}{c|}{34.87} & \multicolumn{1}{c|}{0.0550} & 8.1 & \multicolumn{1}{c|}{32.74} & \multicolumn{1}{c|}{0.0121} & 8.9 & \multicolumn{1}{c|}{33.16} & \multicolumn{1}{c|}{ - } & \multicolumn{1}{c|}{ - } & \multicolumn{1}{c|}{33.65} & \multicolumn{1}{c|}{0.0919} &\multicolumn{1}{c|}{9.5} \\ \hline
(SO,SO) & \multicolumn{1}{c|}{26.41} & \multicolumn{1}{c|}{0.1014} & 7.9 & \multicolumn{1}{c|}{25.26} & \multicolumn{1}{c|}{0.0255} & 8.7 & \multicolumn{1}{c|}{27.23} & \multicolumn{1}{c|}{ - } & \multicolumn{1}{c|}{ - } & \multicolumn{1}{c|}{26.77} & \multicolumn{1}{c|}{ - } &\multicolumn{1}{c|}{ - } \\ \hline
\end{tabular}
\vspace{-3mm}
\caption{Bright Information Metric for images in each brightness category with FMR.}
\vspace{-5mm}
\label{information_of_five_brightness_category}
\end{table*}

\begin{table*}[t]
\centering
\begin{tabular}{|c|c|cc|cc|cc|cc|}
\hline
FSB          & \multirow{2}{*}{Range} & \multicolumn{2}{c|}{C Female}                       & \multicolumn{2}{c|}{C Male}                         & \multicolumn{2}{c|}{A-A Female}            & \multicolumn{2}{c|}{A-A Male}                        \\ \cline{1-1} \cline{3-10} 
Sliding-bins &                        & \multicolumn{1}{c|}{BIM}            & $d'$             & \multicolumn{1}{c|}{BIM}            & $d'$             & \multicolumn{1}{c|}{BIM}            & $d'$    & \multicolumn{1}{c|}{BIM}            & $d'$              \\ \hline
(M6,M6)      & 145 - 185              & \multicolumn{1}{c|}{35.42}          & 8.40          & \multicolumn{1}{c|}{34.38}          & 9.15          & \multicolumn{1}{c|}{36.00}          & 8.36 & \multicolumn{1}{c|}{36.70}          & 9.93           \\ \hline
(M7,M7)      & 150 - 190              & \multicolumn{1}{c|}{35.90}          & 8.47          & \multicolumn{1}{c|}{34.73}          & 9.17          & \multicolumn{1}{c|}{36.22}          & 8.40 & \multicolumn{1}{c|}{36.97}          & 9.94           \\ \hline
(M8,M8)      & 155 - 195              & \multicolumn{1}{c|}{36.30}          & 8.55          & \multicolumn{1}{c|}{34.99}          & 9.20          & \multicolumn{1}{c|}{36.35}          & 8.40 & \multicolumn{1}{c|}{37.09}          & 10.00          \\ \hline
(M9,M9)      & 160 - 200              & \multicolumn{1}{c|}{36.62}          & \textbf{8.59} & \multicolumn{1}{c|}{35.18}          & \textbf{9.20} & \multicolumn{1}{c|}{36.39}          & 8.42 & \multicolumn{1}{c|}{\textbf{37.11}} & \textbf{10.02} \\ \hline
(M10,M10)    & 165 - 205              & \multicolumn{1}{c|}{\textbf{36.85}} & 8.54          & \multicolumn{1}{c|}{\textbf{35.29}} & 9.18          & \multicolumn{1}{c|}{\textbf{36.29}} & 8.44 & \multicolumn{1}{c|}{36.96}          & 9.99           \\ \hline
(M11,M11)    & 170 - 210              & \multicolumn{1}{c|}{36.83}          & 8.47          & \multicolumn{1}{c|}{35.29}          & 9.09          & \multicolumn{1}{c|}{36.01}          & 8.47 & \multicolumn{1}{c|}{36.70}          & 9.97           \\ \hline
(M12,M12)    & 175 - 215              & \multicolumn{1}{c|}{36.66}          & 8.49          & \multicolumn{1}{c|}{35.11}          & 9.01          & \multicolumn{1}{c|}{35.69}          & 8.47 & \multicolumn{1}{c|}{36.34}          & 9.96           \\ \hline
(M13,M13)    & 180 - 220              & \multicolumn{1}{c|}{36.24}          & 8.31          & \multicolumn{1}{c|}{34.63}          & 8.97          & \multicolumn{1}{c|}{35.20}          & 8.51 & \multicolumn{1}{c|}{35.76}          & 9.88           \\ \hline

\end{tabular}
\caption{Average BIM and d-prime of each sliding-bin.}
\vspace{-3mm}
\label{information_of_sliding-bins}
\end{table*}
FMR variation across demographics in our results mostly follow the pattern seen in the literature
\cite{FRVT_2019_Part3, krishnapriya2020issues}.
Considering (M,M) brightness pairs, the Caucasian female ArcFace FMR is a factor of 3.7 times that for Caucasian male, and the COTS-D FMR is a factor of 5.5 times that for Caucasian male.
For African-American male, the ArcFace FMR is a factor of 3.9 that for Caucasian male, but the COTS-D FMR is 65\% {\it lower} than that for Caucasian male.
For African-American female, the ArcFace FMR is a factor of 34 times that for Caucasian male, and the COTS-D FMR is a factor of 2.7 times that for Caucasian male.
For ArcFace, using a fixed similarity threshold to generate a FMR results in Caucasian male having the lowest FMR, similar and higher FMRs for Caucasian female and African-American male, and much higher FMR for African-American female.
We will comment on the relative d-prime separation of impostor and genuine scores later.

The relative COTS-D FMR for Caucasian male versus African-American male is clearly at odds with what is seen in the literature.
The black-box nature of a COTS algorithm means that we cannot offer any confident speculation of the cause for this.
But we observe that there is no overlap between MORPH and the dataset used in training ArcFace, whereas the training and development data for COTS-D is unknown to us.
We also observe that the COTS-D impostor and genuine distributions are distinctly not Gaussian.
Similarity scores from COTS-D are in the range from 0 to 1.
Across the four demographics, from 58\% to 80\% of the impostor scores are reported as 0, and 98\% or more of the genuine scores are reported as 1. This suggests a thresholding step on the raw similarity scores before they are reported. 
As a result, it is meaningless to plot impostor and genuine distributions or to report d-prime values for COTS-D.

General conclusions from this analysis are as follows.
One, there is target range in the middle of the observed brightness distribution that results in lowest FMR.
Two, image pairs with similar brightness, darker or brighter than the target range, result in increased FMR.
Three, image pairs with substantially different brightness  result in decreased FMR.
Thus the current standard approach in which the same matching threshold is used for all image pairs,
results in some pairs having predictably higher and lower FMR based on their brightness.
For controlled image acquisition, this problem can potentially be addressed by adjusting lighting to acquire images in the target range of face skin brightness.

\section{Brightness, Face Information and d-prime}
Brightness metrics in previous work \cite{4168417, long2011near, nasrollahi2009complete} suggest that an image with higher brightness is of higher quality. 
(See Related Work.)
However, the previous section shows that the FMR of similar-brightness impostor pairs is lowest for middle brightness and increases for pairs that are too dark or too bright.
This section explores how the level of variability in face skin brightness can explain the accuracy across brightness categories.
We propose a Brightness Information Metric (BIM) that captures the degree of variation in face skin brightness, defined as follows:
\vspace{-3mm}
\begin{equation}
    BIM = \sum_{i=0}^{N}|B_{i} - \Bar{B}|P(B_{i})
\vspace{-3mm}
\end{equation}
where $B_{i}$ is a brightness level, $P(B_{i})$ is the probability of face skin pixels that have this brightness level, and $\bar{B}$ is the average brightness of the face skin pixels. 
Note that, since the brightness values distribute in a Gaussian-like form, the BIM is positively related to the average information entropy of the selected group of images.

Table~\ref{information_of_five_brightness_category} summarizes the average BIM across the different brightness categories and demographics, along with FMR and the d-prime for the separation between impostor and genuine distributions.
ArcFace d-prime and FMR values are shown because, as explained earlier, the COTS-D impostor and genuine distributions contain large spikes at zero and one, respectively.
For each of the four demographics, the largest BIM and the lowest FMR occur for M brightness. 
For all demographics, the BIM decreases from (M, M) to (O,O) and decreases again to (SO,SO), and it also decreases from (M,M) to (U,U) and decreases again to (SU,SU).
Across brightness categories and demographics, a lower BIM for images in an impostor pair means less information to distinguish between individuals, and so a higher similarity score and higher FMR.

\subsection{Target Brightness Range for Best Accuracy}
\label{target_brightness}

The M brightness range in earlier sections 
corresponds to the middle 70\% of the overall brightness distribution.
Our results so far suggest it should be possible to select a narrower range that gives better accuracy.
To investigate the feasibility and generality of this, we conduct a refined analysis of our M brightness range.

Table~\ref{information_of_sliding-bins} summarizes  BIM and d-prime for image pairs in sliding 40-brightness-level bands of our initial M brightness range. Across all four demographics, the highest BIM and d-prime are generally found for M9 and M10, with combined brightness range 160-205.
The operational scenario brightness range in Figure~\ref{fig:brightness_distributions} runs about 50 to 250, meaning that the target range for  best accuracy is about one-fourth the observed operational range.
Also, note that 28\% of  African-American female images, 24\% of  African-American male images, 62\% of Caucasian male images and 59\% of Caucasian female images in MORPH fall into the target range for best accuracy. 
Thus it appears that a portion of the demographic variation in accuracy is due to variation in quality of the images acquired.

\section{Conclusions and Discussion}
Using a face image dataset from a real operational scenario with controlled image acquisition, we show that the distribution of face skin brightness varies over most of the 0-255 range.
While the distribution of face skin brightness has large overlap across demographics, on average, images of African-Americans have lower brightness than images of Caucasians and images of males have (slightly) lower brightness than images of females. Also, the standard deviation of the brightness metric is about one-third greater for African-Americans than for Caucasians.

To analyze how accuracy varies with image pair brightness, the face skin brightness distribution was divided into intervals representing strongly under-exposed, under-exposed, middle exposure, over-exposed and strongly over-exposed.
FMR analysis from two different matchers showed very similar trends.
Pairs of images that have lower or higher face skin brightness have an increased FMR relative to image pairs with middle brightness.
Also, pairs of images that differ strongly in face skin brightness have a lower FMR than image pairs with middle brightness.
We used a brightness information metric that measures the variation in brightness in the face skin region to explain that the available information in the face skin region is at a maximum at a middle brightness level and declines as the face skin brightness moves toward either end of the scale.

We showed that it is possible to find a target face skin brightness range that represents the maximum BIM for all four demographics, and achieves the maximum d-prime separation of impostor and genuine distributions.
This target brightness range is about one-fourth the range of brightness seen in a retrospective image dataset from an operational scenario with controlled image acquisition.
This analysis suggests that operational scenarios with controlled image acquisition could maximize face recognition accuracy by adjusting the lighting on image acquisition to hit the target face skin brightness range.
The analogy could be made here to image acquisition for commercial iris recognition, in which the illumination is controlled and image quality is automatically checked at the time of acquisition.

Analysis suggests that the issue of acquiring images with face skin brightness in an appropriate range for all demographics is related to the problem of varying accuracy across demographics.
Consider that our analysis suggests a target face skin brightness range of 160-205.
The face skin brightness distribution shows that different demographics have a very different fraction of their images falling into this range.
Caucasian males and females have 62\% and 59\% of their images falling into this interval.
But African-American males and females have just 24\% and 28\% of their images falling into this interval.
Acquiring images of equal quality for all demographics may be an essential element to expecting equal accuracy across demographics.
{\small
\bibliographystyle{ieee_fullname}
\bibliography{egbib}
}

\end{document}


\title{Response to Reviews on ``Face Recognition Accuracy Across Demographics
:\\Shining a Light Into the Problem''
} %

\maketitle

Thanks for the suggestions and comments. Review comments are in red and responses
 in black.

Some reviews suggest lack of novelty because ``Brightness and face image quality matter when it comes to face recognition is well-known''.
Illumination in general is of course well studied, but this paper focuses on
the role of brightness in unequal recognition accuracy across demographics, which has {\bf not} been studied before.

\noindent
{\bf Review \#1.}\\
\textbf{\textcolor{red}{Brightness and face image quality matter when it comes to face recognition, which is well known ...}}

\noindent
\textit{Pose and illumination have been analyzed for decades. But current face recognition models show unequal recognition accuracy across demographic groups, which is the focus of this paper. The role of brightness in this is not well-known.}

\noindent
\textbf{\textcolor{red}{The claim to work for all demographic variations seems not correct, since the demographic groups selected are not really diverse.}}

\noindent
\textit{We mean ``all demographics'' to be all demographics analyzed in this paper, male/female and Caucasian/African-American. We will clarify it further.}

\noindent
{\bf Review \#2.}\\
\textbf{\textcolor{red}{The authors have not properly reviewed the mainstream or dominant approaches developed in computer vision for illumination invariant face/object recognition.}}

\noindent
\textit{
We are not repeating work on illumination-invariant recognition in general. We are evaluating how illumination in an operational scenario impacts accuracy differently across demographics, and what can be done about it.}

\noindent
\textbf{\textcolor{red}{The FSB is an inadequate and unreliable measure of face surface brightness. For example, if one uses an illumination dimmer, the pixel values of an image are shifted by some constant either up or down. Those types of images would have two different FSB measures, despite an inconsequential difference.}}

\noindent
\textit{The reviewer is trying to argue that mis-measurement on brightness can cause the mis-measurement of quality. Our quality measurement metric BIM is invariant in this case.}

\noindent
\textbf{\textcolor{red}{The authors also propose a brightness information measure (BIM) that is computed by multiply $|B\_i - mean (B)|$ with $P(B_i)$. Why not $P(|B_i - mean(B)|)?$}}

\noindent
Both equations can reflect variance of face skin brightness.

\noindent
\textbf{\textcolor{red}{Both face verification (analysis of pairs of images) and face classification (the more general face recognition problem) are impacted by image exposure. Solutions for addressing over-exposure and under-exposure problems have been implemented both in software and hardware with high-dynamic range imaging techniques. ...
}}
 
\noindent
\textit{This paper focuses on understanding and mitigating unequal accuracy across demographics. 
This review seems to think that this is already solved.
The 4 papers cited either don't have experiments on how to improve recognition accuracy, or the models/datasets used are not state-of-the-art.}\\

\noindent {\bf Review \#3.}\\
\textbf{\textcolor{red}{Limited algorithmic novelty}}
 
\noindent
\textit{Our contribution focuses on better understanding of the causes of accuracy difference across demographics.
This is an important current topic.
}

\noindent
\textbf{\textcolor{red}{The importance of FSB is not clear when similar conclusions can be drawn using iFace or any other ...
}}

\noindent
\textit{We show that FSB is better than the commercial brightness metric. 
We want to use a better tool if we have one available, to make our experiments as rigorous as possible..
}

\noindent
\textbf{\textcolor{red}{BIM measures the variability of brightness across a category. Isn't lower better?}}
 
\noindent
\textit{From our observation, strongly under-exposed and strongly over-exposed images has much lower variability on brightness (see samples in Figure 4), which affects the recognition accuracy a lot on some demographic groups. Therefore, the higher BIM value means better.}\\

\noindent
{\bf Review \#4.}\\
\noindent
\textbf{\textcolor{red}{Evaluating the brightness of an image (face region) is not a novel idea, and it has been studied for decades, where the proposed approach is a simple average of all skin-pixels. Lacking diversity in large public datasets is already a well-known problem.}}

\noindent
\textit{Our proposed brightness metric is better than the commercial iFace SDK, and is a tool used to investigate how brightness affects accuracy across demographic groups. The point of the paper is investigating accuracy across demographics.
}

\noindent
\textbf{\textcolor{red}{Not all ML/CV models are vulnerable to skin brightness, and the study only evaluated a few models and metrics.}}
 
\noindent
\textit{We used ArcFace and a recent commercial SDK. These are a reasonable representation of the state-of-the-art.
}

\noindent
\textbf{\textcolor{red}{Applying the proposed metric for a pair of images has limited scope.}}
 
\noindent
\textit{This research is focusing on images captured in a stable acquisition environment, and the FSB and BIM metrics work well for getting an accurate brightness value and reflecting the variance of brightness (See in Figure 3 and Table 2).}